\documentclass[acmtog]{acmart}

\usepackage{booktabs} 
\usepackage{enumitem}
\usepackage{makecell} 
\usepackage{multirow} 
\usepackage{xcolor} 
\usepackage{marvosym}

\renewcommand\footnotetextcopyrightpermission[1]{} 
\usepackage[ruled]{algorithm2e} 

\SetAlFnt{\small}
\SetAlCapFnt{\small}
\SetAlCapNameFnt{\small}
\SetAlCapHSkip{0pt}

\acmJournal{TOG}

\copyrightyear{2025}
\acmYear{2025}
\setcopyright{cc}
\setcctype{by}
\acmConference[SIGGRAPH Conference Papers '25]{Special Interest Group on Computer Graphics and Interactive Techniques Conference Conference Papers }{August 10--14, 2025}{Vancouver, BC, Canada}
\acmBooktitle{Special Interest Group on Computer Graphics and Interactive Techniques Conference Conference Papers (SIGGRAPH Conference Papers '25), August 10--14, 2025, Vancouver, BC, Canada}
\acmDOI{10.1145/3721238.3730683}
\acmISBN{979-8-4007-1540-2/2025/08}

\begin{document}

\title{FlexiAct: Towards Flexible Action Control in Heterogeneous Scenarios}

\author{Shiyi Zhang*}
\affiliation{%
 \institution{Tsinghua Shenzhen International Graduate School, Tsinghua University}
 \country{China}}
\email{sy-zhang23@mails.tsinghua.edu.cn}
\author{Junhao Zhuang}
\authornote{Equal contribution}
\affiliation{%
 \institution{Tsinghua Shenzhen International Graduate School, Tsinghua University}
 \country{China}}
\email{zhuangjh23@mails.tsinghua.edu.cn}
\author{Zhaoyang Zhang}
\authornote{Project lead: Zhaoyang Zhang (zhaoyangzhang@link.cuhk.edu.hk)}
\affiliation{%
 \institution{Tencent ARC Lab}
 \country{China}}
\email{zhaoyangzhang@link.cuhk.edu.hk}
\author{Ying Shan}
\affiliation{%
 \institution{Tencent ARC Lab}
 \country{China}}
\email{yingsshan@tencent.com}
\author{Yansong Tang}
\authornote{Corresponding author: Yansong Tang (tang.yansong@sz.tsinghua.edu.cn)}
\affiliation{%
 \institution{Tsinghua Shenzhen International Graduate School, Tsinghua University}
 \country{China}}
\email{tang.yansong@sz.tsinghua.edu.cn}

\newcommand{\red}[1]{\textcolor{black}{#1}}
\newcommand{\blue}[1]{\textcolor{black}{#1}}
\newcommand{\method}{FlexiAct\xspace} 
\newcommand{\stageone}{RefAdapter\xspace} 
\newcommand{\stagetwo}{FAE\xspace} 

\begin{abstract}
Action customization involves generating videos where the subject performs actions dictated by input control signals. Current methods use pose-guided or global motion customization but are limited by strict constraints on spatial structure such as layout, skeleton, and viewpoint consistency, reducing adaptability across diverse subjects and scenarios.
To overcome these limitations, we propose \method, which transfers actions from a reference video to an arbitrary target image. Unlike existing methods, \method allows for variations in layout, viewpoint, and skeletal structure between the subject of the reference video and the target image, while maintaining identity consistency.
Achieving this requires precise action control, spatial structure adaptation, and consistency preservation. To this end, we introduce \stageone, a lightweight image-conditioned adapter that excels in spatial adaptation and consistency preservation, surpassing existing methods in balancing appearance consistency and structural flexibility.
Additionally, based on our observations, the denoising process exhibits varying levels of attention to motion (low frequency) and appearance details (high frequency) at different timesteps. So we propose \stagetwo (Frequency-aware Action Extraction), which, unlike existing methods that rely on separate spatial-temporal architectures, directly achieves action extraction during the denoising process.
Experiments demonstrate that our method effectively transfers actions to subjects with diverse layouts, skeletons, and viewpoints. We release our code and model weights to support further research at \href{https://shiyi-zh0408.github.io/projectpages/FlexiAct/}{\textcolor{red}{FlexiAct}}.
\vspace{-5pt}

\end{abstract}

\begin{teaserfigure}
\centering
  \vspace{-8pt}
  \includegraphics[width=\linewidth]{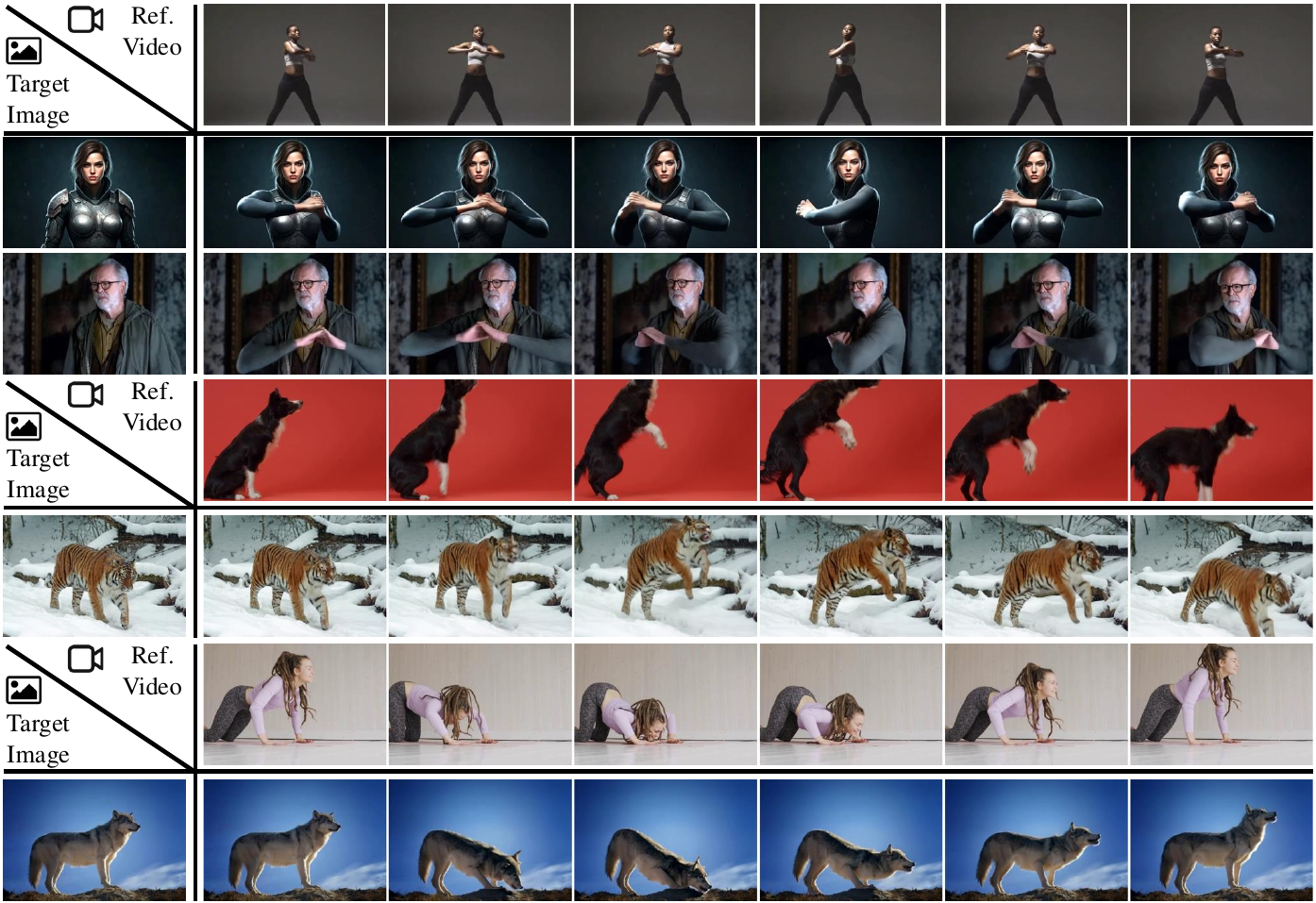}
  \vspace{-20pt}
  \caption{\textbf{Visualization for our \method results.} Given a target image, FlexiAct transfers actions from a reference video to the target subject, achieving accurate motion adaptation and appearance consistency even in heterogeneous scenarios with varying spatial structures or cross-domain subjects.}
  \vspace{-2pt}
  \label{fig:teaser}
\end{teaserfigure}

%
%



\maketitle

\vspace{-8pt}
\section{Introduction}
\begin{figure*}
    \centering
    \includegraphics[width=\textwidth]{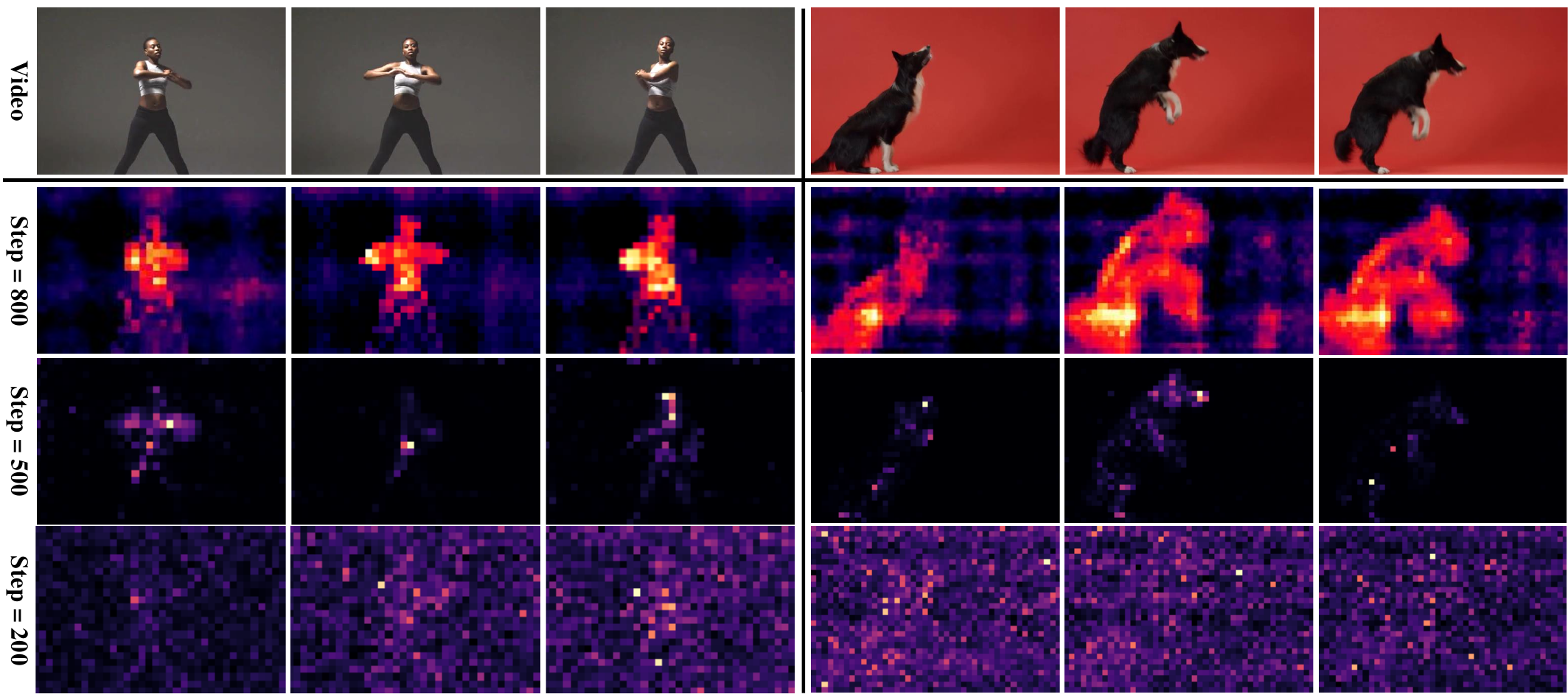}
    \vspace{-20pt}
    \caption{\textbf{Attention maps between our frequency-aware embeddings and video tokens in the MMDiT at different denoising timesteps.} Our embeddings focus on low-frequency motion information (e.g., motion regions) in early denoising stages and shift to high-frequency details in later stages.}
    \vspace{-12pt}
    \label{fig:attn_map}
\end{figure*}

Action transfer involves applying specific actions to a target subject and is widely used in films, games, and animation. However, it often requires substantial financial and human resources. For example, professional motion capture systems can cost tens of thousands of dollars and require skilled technicians. Similarly, creating a 30-second animation at 12 frames per second can take about 20 work days from six professional animators. These high costs pose significant challenges, limiting access for many potential creators.

In response to these limitations, significant efforts have been devoted to achieving motion control in video generation, which can be broadly categorized into two main approaches:
(1) \textbf{Predefined signals} methods using signals like pose and depth maps, such as AnymateAnyone~\cite{hu2024animate} and StableAnimator~\cite{tu2024stableanimator}, and (2) \textbf{Global motion} methods like Motion Director\cite{zhao2023motiondirector} and Motion Inversion~\cite{wang2024motion}. Despite advancements, these methods exhibit notable limitations. Predefined signal methods require strict alignment of spatial structures (e.g., shape, skeleton, viewpoint) between the target image and reference video, which is often not feasible in real-world scenarios. They also struggle with obtaining pose information for non-human subjects. 
Conversely, global motion methods typically generate motions with fixed layouts and cannot transfer motion across diverse subjects. Some approaches~\cite{zhao2023motiondirector} employ identity-specific Low-Rank Adaptations (LoRAs)~\cite{hu2021lora} for animation, yet they encounter difficulties with appearance consistency and flexibility.

\red{To this end, we introduce \method, an Image-to-Video (I2V) framework for flexible action customization in heterogeneous scenarios. Given a reference video and an arbitrary target image, our method transfers actions from the reference video to the target image without alignment in layout, shape, or viewpoint, preserving both action dynamics and appearance details. \method builds upon CogVideoX-I2V \cite{yang2024cogvideox} with a two-stage training on two novel components: RefAdapter and Frequency-aware Action Extraction (FAE), addressing the following challenges:
(1) \textbf{Spatial structure adaptation:} Adapting actions to target images with different poses, layouts, or viewpoints.
(2) \textbf{Precise action extraction and control:} Accurately decoupling and replicating action from the reference video.}

\red{RefAdapter addresses the first challenge, which is an image-conditioned architecture generating videos given the input images. It combines the accuracy of I2V frameworks with the flexibility of conditional injection architectures like IP-Adapter \cite{ye2023ip-adapter}, ensuring appearance consistency between the video and the conditioning image while avoiding strict constraints on the first video frame. This enables FlexiAct to adapt reference motion to various spatial structures using arbitrary frames as image conditions. RefAdapter requires only low training costs, finetuning a small set of LoRA, avoiding the large parameter replication in ReferenceNet \cite{hu2024animate} and ControlNet \cite{zhang2023adding}.}

\red{For precise action control, we propose Frequency-aware Action Extraction (\stagetwo). This method incorporates a set of learnable embeddings to capture entangled video information from the reference video during training. As illustrated in Figure \ref{fig:attn_map}, we observe that these embeddings dynamically adjust their attention to different frequency components across denoising timesteps. Specifically, they prioritize motion information (low-frequency features) in early timesteps and shift focus to appearance details (high-frequency features) in later timesteps. Leveraging this property, \stagetwo performs action extraction directly during the denoising process by modulating attention weights at different timesteps, eliminating the need for separate spatial-temporal architectures.}

To validate the effectiveness of \method, we establish a benchmark for heterogeneous scenarios. Experiments demonstrate \method's flexible and general action transfer capabilities. As shown in Figure \ref{fig:teaser}, \method accurately transfers action from a reference video to subjects with varying layouts, viewpoints, shapes, and even domains, while maintaining appearance consistency. 

In summary, our paper makes the following key contributions:
\vspace{-4pt}
\begin{itemize}[itemsep=0pt,leftmargin=*]
\item We propose \method, a flexible action transfer method that \textbf{first} adapts reference actions to arbitrary subjects with diverse spatial structures while ensuring action and appearance consistency.
\item We introduce RefAdapter, which achieves spatial structure adaptation and appearance consistency with a few trainable parameters.
\item We propose Frequency-aware Action Extraction, which precisely extracts action and controls the video synthesis during sampling.
\item Our extensive experiments demonstrate \method's capabilities across diverse scenarios, including various subjects and domains.
\end{itemize}
\vspace{-14pt}

\section{Related Work}

\subsection{Global Motion Customization}
Global Motion Customization focuses on transferring the overall motion dynamics from a reference video, such as camera movements, object trajectories, and actions, to generate videos with consistent global motion patterns~\cite{yatim2023space, jeong2023vmc, zhao2023motiondirector, ling2024motionclone, jeong2024dreammotion}. 
The challenge of this task lies in effectively extracting motion from the reference video.
Recent work like Motion Director\cite{zhao2023motiondirector} addresses this by adopting spatial-temporal LoRA\cite{hu2021lora} to decouple the appearance and the motion.
Meanwhile, Diffusion Motion Transfer ~\cite{yatim2023space} extracts motion via a handcrafted loss during inference. 
On the other hand, Video Motion Customization ~\cite{jeong2023vmc} encodes motion directly into the text-to-video model. 
Motion Inversion~\cite{wang2024motion} introduces two types of embeddings to decouple the appearance and motion. 
However, most of these methods fail to adapt motion to specific subjects, as they primarily focus on generating videos that approximate the layout of the reference video. 
In contrast, we propose a framework that can handle the action transfer in heterogeneous scenarios. Furthermore, inspired by the insights into noise schedules from~\cite{si2024freeu,lin2024common,wu2023freeinit,lu2024freelong,qiu2023freenoise} and our observations regarding different denoising timesteps, we propose the first denoising process-based action extraction framework.

\vspace{-12pt}
\subsection{Predefined signal-based Action Customization}

Action customization methods based on predefined signals, such as pose, depth, and edges, transfer motion from these signals to animate target images~\cite{he2022lvdm, esser2023structure, wang2023videocomposer}. They focus on how to precisely control subjects with identical spatial structures using predefined signals. Early approaches~\cite{Siarohin_2019_NeurIPS, siarohin2021motion, huang2021few} primarily utilized GANs~\cite{goodfellow2020generative} for reference animation. Recent advancements have shifted to diffusion models, with Disco~\cite{wang2024disco} pioneering this transition for image animation. Subsequent works, such as MagicAnimate~\cite{xu2024magicanimate} and AnimateAnyone~\cite{hu2024animate}, employ ReferenceNet and pose nets to decouple pose and appearance modeling. Further innovations include Champ~\cite{zhu2024champ}, which integrates 3D SMPL signals for enhanced controllability, and Unianimate~\cite{wang2024unianimate}, which incorporates Mamba~\cite{mamba2} into diffusion models for improved efficiency. Additionally, MimicMotion~\cite{zhang2024mimicmotion} introduces a regional loss to mitigate distortion, while ControlNeXt~\cite{peng2024controlnext} replaces the computationally intensive ControlNet~\cite{zhang2023adding} with a lightweight convolution-based pose net. 
However, these methods remain heavily reliant on predefined signals, limiting their effectiveness when the target image and reference video exhibit significant spatial discrepancies, such as variations in shape or pose. Moreover, they face challenges in non-human scenarios, where predefined motion signals are often unavailable or difficult to obtain. 
In contrast, our method does not rely on predefined signals with numerous constraints, enabling it to handle more general scenarios, such as transferring actions between subjects with different shapes, skeletons, viewpoints, and even across domains.

\vspace{-12pt}
\subsection{Customized Video Generation via Condition Injection}

With the advancement of text-to-video models~\cite{esser2023structure, zhang2024show,yuan2024instructvideo,bar2024lumiere, guo2023animatediff, blattmann2023align, chen2023videocrafter1,chen2024videocrafter2, wang2023modelscope, wang2023lavie, wang2023videofactory, videoworldsimulators2024, yang2024cogvideox, ma2024latte, zhou2024allegro}, customized video generation has emerged as a critical and highly active research topic. 
Among these methods, some focus on injecting control signals into the video generation process through condition injection, which can generally be categorized into two types: one based on cross-attention injection, such as IP-Adapter\cite{ye2023ip}, and the other on module duplication for layer-wise injection, such as ReferenceNet\cite{hu2024animate}. Cross-attention approaches, though lightweight, often fail to ensure appearance consistency due to coarse-grained representations (e.g., CLIP image features~\cite{radford2021learning}). Module duplication enables finer control but incurs high training costs from parameter replication. 
\blue{In contrast, based on the I2V model, our RefAdapter strikes a balance, achieving ReferenceNet-level fine-grained appearance control while requiring only a small number of training parameters. Additionally, RefAdapter can reduce the strict first-frame dependency of I2V models.}

\begin{figure*}
    \centering
    \includegraphics[width=\textwidth]{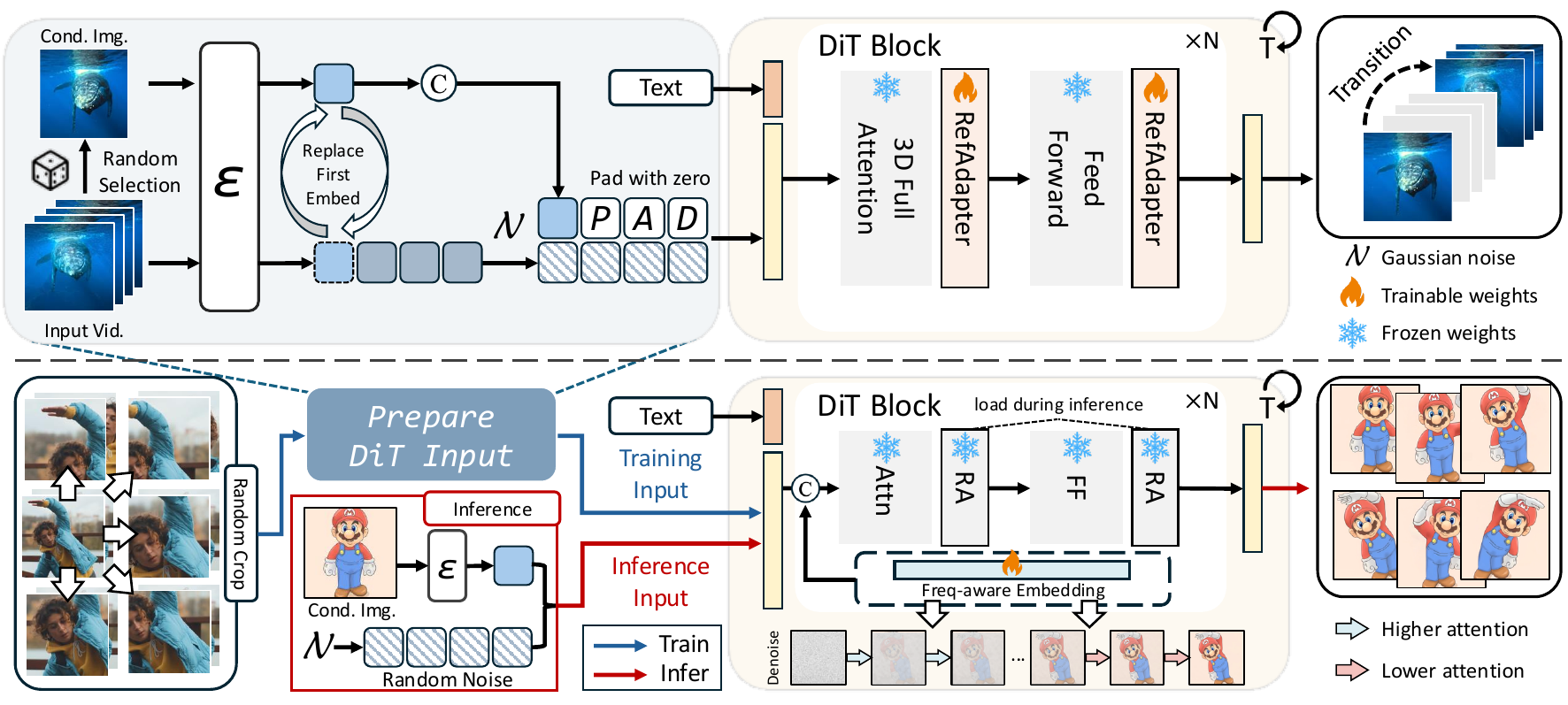}
    \vspace{-12pt}
    \caption{\textbf{Overview of FlexiAct.} (1) The upper part illustrates RefAdapter's training, which conditions arbitrary frames to enable transitions across diverse spatial structures. (2) The lower part outlines \stagetwo's training and inference, where attention weights of video tokens to the frequency-aware embedding are dynamically adjusted based on timesteps, facilitating action extraction.}
    \vspace{-8pt}
    \label{fig:method}
\end{figure*}
\vspace{-18pt}
\section{method}
\subsection{Overview}
As illustrated in Figure \ref{fig:method}, \method builds upon CogVideoX-I2V with a two-stage training on two components: RefAdapter and Frequency-aware Action Extraction (FAE). RefAdapter facilitates action adaptation to subjects with varying spatial structures while maintaining appearance consistency.
FAE dynamically adjusts attention weights to frequency-aware embeddings at different denoising timesteps, enabling effective action extraction. These two components are trained separately to ensure that action extraction does not interfere with RefAdapter's consistency preservation. Section \ref{subsec:cog} introduces FlexiAct's base model. Section \ref{subsec:refadapter} introduces RefAdapter and its training methodology. Section \ref{subsec:tae} details the training and inference processes of \stagetwo. Finally, Section \ref{subsec:pipeline} outlines the training and inference pipeline of FlexiAct.
\vspace{-8pt}
\red{\subsection{Basis Image-to-Video Diffusion Model}}
\label{subsec:cog}

\red{We use CogVideoX-I2V \cite{yang2024cogvideox} as our basis image-to-video (I2V) model. CogVideoX-I2V is an MMDiT-based \cite{esser2024scaling} video diffusion model that operates in a latent space. Given an image \(\mathbf{I}\in\mathbb{R}^{H\times W\times3}\) and a textual prompt, CogVideoX-I2V generates a video \(\mathbf{V} \in \mathbb{R}^{T \times H \times W \times 3}\). CogVideoX-I2V utilizes a 3D VAE to map condition images and videos into the latent space. For video inputs, the 3D VAE encoder (\(\epsilon\) in Figure \ref{fig:method}) compresses both temporal and spatial dimension, producing a latent \(L_{video}\in\mathbb{R}^{\frac{T}{4}\times\frac{H}{8}\times\frac{W}{8}\times C}\), where \(C\) denotes the channel number. For image inputs (\(T=1\)), the encoder preserves the temporal dimension, yielding \(L_{image}\in\mathbb{R}^{1\times\frac{H}{8}\times\frac{W}{8}\times C}\), which is then zero-padded along the temporal dimension to match \(L_{video}\)'s shape (\(1\rightarrow\frac{T}{4}\)). During inference, this padded \(L_{image}\) is concatenated with a random noise \(\mathcal{N}\in\mathbb{R}^{\frac{T}{4}\times\frac{H}{8}\times\frac{W}{8}\times C}\) along the channel dimension for subsequent denoising process. During training, the first frame of the ground truth video serves as the input image; its padded \(L_{image}\) is concatenated with the noisy \(L_{video}\) along the channel dimension to predict the added noise. This process is optimized with MSE loss between the added and predicted noise, consistent with classic diffusion models \cite{ho2020denoising}.}
\vspace{-8pt}
\subsection{RefAdapter}
\label{subsec:refadapter}
The upper part of Figure \ref{fig:method} illustrates the RefAdapter training process. We note that directly using I2V for spatial structure adaptation is challenging because: (1) the action extraction process compromises the I2V model's consistency preservation, and (2) I2V is a strongly constrained image-conditioned framework. 
During training, I2V uses the first video frame as the condition image, ensuring video consistency but potentially hindering smooth action transfer if the initial spatial structure differs from the reference video. 

\red{To address this, we introduce a gap between the condition image and the initial spatial structure by using a randomly sampled frame from the unedited video as the condition image during training. Specifically, we propose \textbf{RefAdapter}, which includes LoRA injected into CogVideoX-I2V's MMDiT layers. RefAdapter is trained on 42,000 videos from \cite{ju2024miradata} in a 40,000-step one-time training. The training process of RefAdapter mostly follows CogVideoX-I2V, with key distinctions: (1) The condition image is randomly selected from the entire untrimmed video instead of the first frame, maximizing spatial structure discrepancy. (2) We replace the first embedding along the temporal dimension of \(L_{video}\) with \(L_{image}\), enabling the model to use the first embedding as a reference for guiding video generation, rather than constraining it as the video's starting point. Without this replacement, the generated video's initial state would remain constrained to match the condition image.}

\vspace{-8pt}
\subsection{Frequency-aware Action Extraction}
\label{subsec:tae}

To extract action information from a reference video, a straightforward approach involves training motion embeddings to fit motion information, akin to Genie~\cite{bruce2024genie} or textual inversion~\cite{gal2022textual}. However, our initial attempts with this method yield suboptimal results. Inspired by~\cite{si2024freeu,lin2024common,wu2023freeinit,lu2024freelong,qiu2023freenoise}, we delve into the relationship between the embeddings and action information during the denoising process. By examining the attention maps between motion embeddings and video tokens across different denoising timesteps, as visualized in Figure~\ref{fig:attn_map}, we observe that our embeddings predominantly focus on low-frequency action information in the early stages, gradually shifting their attention to high-frequency details in the later stages. Leveraging this insight, we propose the Frequency-aware Action Extraction.

\red{Specifically, the lower part of Figure \ref{fig:method} illustrates the training and inference process of \stagetwo. We train \textbf{Frequency-aware Embedding} for individual reference videos, which includes learnable parameters concatenated to MMDiT layers' inputs. This training differs from CogVideoX-I2V by applying random cropping on the input video to prevent the Frequency-aware Embedding from focusing on the reference video's layout. RefAdapter is not loaded during this training to protect its conditioning ability. After training, the Frequency-aware Embedding captures both motion and appearance from the reference video.}

During inference, \stagetwo extracts action information and adapts it to the target image. As shown in Figure \ref{fig:attn_map}, attention maps between frequency-aware embeddings and video tokens reveal that at larger timesteps (e.g., step=800), the embeddings focus on motion information (low frequency), with high attention on the moving parts of the subject. At intermediate timesteps (e.g., step=500), the focus shifts to fine-grained details of the subject. At later timesteps (e.g., step=200), attention is distributed across the entire image, indicating a focus on global details like the background.

Based on this observation, during inference, we increase the attention weight of video tokens on the frequency-aware embeddings at larger timesteps while maintaining the original weights at other timesteps. This enhances the generated video's ability to perceive and replicate the motion of the reference video. The reweighting strategy of \stagetwo can be formulated as:
\vspace{-5pt}
\begin{equation}
W_{bias} = 
\begin{cases} 
\alpha, & t_{l} \leq t \leq T\\
\dfrac{\alpha}{2} \left[ \cos \left( \dfrac{\pi}{t_{h} - t_{l}} (x - t_{l}) \right) + 1 \right], & t_{h} \leq t < t_{l} \\ 
0, & 0 \leq t < t_{h}
\end{cases}
\end{equation}
\vspace{-5pt}

where \( W_{\text{bias}} \) denotes the bias value applied to the original attention weight. $t$ denotes denoising timestep. The parameter \( \alpha \) controls the strength of the bias, while \( t_l \) and \( t_h \) represent the low-frequency and high-frequency timesteps, respectively. A transition function is employed between these timesteps to smoothly vary \( W_{\text{bias}} \) from low-frequency to high-frequency timesteps. During inference, the attention weight of video tokens to the frequency-aware embedding is dynamically adjusted as \( W_{\text{attn}} = W_{ori} + W_{\text{bias}} \) in all DiT layers. 

Experimental results demonstrate that the attention reweighting strategy improves the generated video's ability to reproduce the reference action. In practice, we set \(\alpha = 1\), \(t_h = 700\), and \(t_l = 800\), and demonstrate the impact of the transition function on the generation results in Figure \ref{fig:supp_ablation}. Without the transition function, changing the bias at \( t = 700 \) would cause the model to learn appearance information from the reference video (e.g., the clothing in Figure \ref{fig:supp_ablation}) while altering the bias at \( t = 800 \) would result in inaccurate motion. This indicates that a transition process between \( t = 700 \) and \( t = 800 \) is necessary to achieve a balance between appearance and motion.
\subsection{Training and Inference Pipeline of FlexiAct}
\label{subsec:pipeline}
\red{As shown in Figure \ref{fig:method}, we first train RefAdapter on a broader dataset (upper part). Subsequently, we train the frequency-aware embedding based on the individual reference video. RefAdapter does not participate in this training process. During the Inference phase, the RefAdapter is loaded, and an arbitrary target image is provided. FAE dynamically adjusts the generated video's attention to Frequency-aware Embedding according to denoising timesteps, transferring actions from the reference video to the target image.}
\section{experiment}
\begin{figure}
    \centering
    \includegraphics[width=\linewidth]{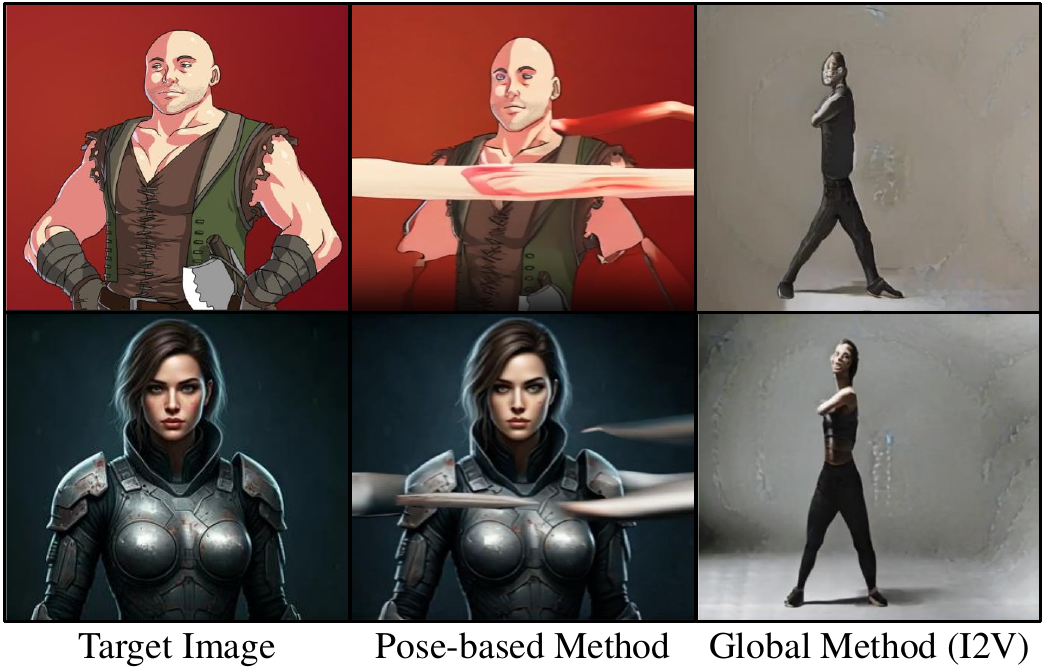}
    \vspace{-18pt}
    \caption{Results of transferring ``turning'' action to the target image using the pose-based method and the animation version of the global motion method.}
    \vspace{-22pt}
    \label{fig:othermethods}
\end{figure}
\begin{figure*}
    \centering
    \includegraphics[width=\textwidth]{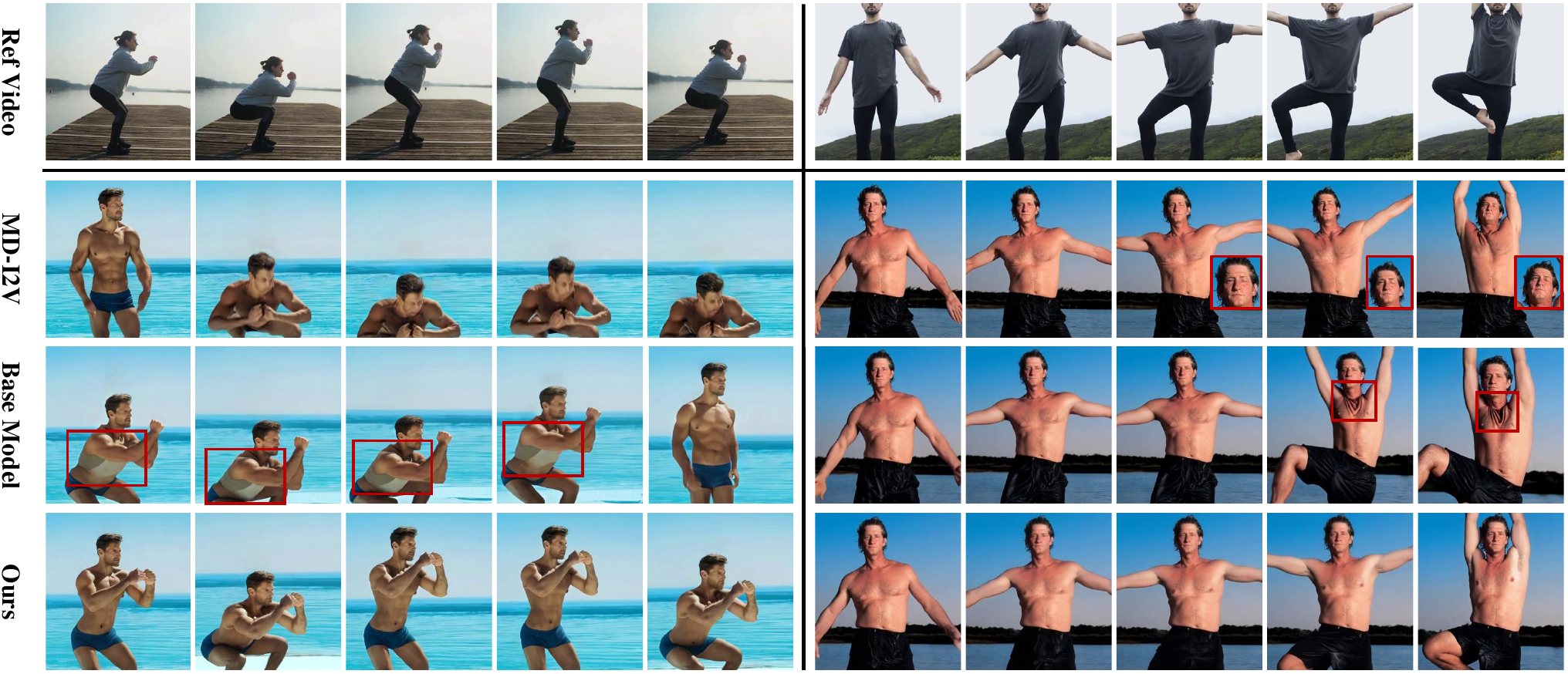}
    \vspace{-20pt}
    \caption{Qualitative comparison of action transfer from reference video (Ref Video) to target images with varying spatial structures. Red boxes highlight regions where the appearance deviates from the target image. Our method demonstrates superior performance in maintaining appearance consistency with the target image and motion fidelity to the reference video compared to other approaches.}
    \label{fig:comparison}
    \vspace{-10pt}
\end{figure*}
\begin{table*}[t]{
    \centering
    \setlength{\tabcolsep}{4pt} 
    \begin{tabular}{c|cccc|ccc}
    \toprule[1.5pt]
        \textbf{\multirow{3}{*}{Method}} & \multicolumn{4}{c|}{Automatic Evaluations} & \multicolumn{3}{c}{Human Evaluations}\\ 
        \cmidrule(lr){2-8}
         & \textbf{\makecell{Text \\Similarity} $\uparrow$} & \textbf{\makecell{Motion \\Fidelity} $\uparrow$}  &\textbf{\makecell{Temporal \\Consistency}$\uparrow$}&\textbf{\makecell{Appearance \\Consistency} $\uparrow$}& & \textbf{\makecell{Motion \\Consistency}}& \textbf{\makecell{Appearance \\Consistency}}\\ 
        \toprule
        MD-I2V~\citep{zhao2023motiondirector}& 0.2446& 0.3496& 0.9276&0.8963&v.s. Base Model & 47.2 v.s. 52.8 & 53.1 v.s. 46.9 \\ 
        Base Model& 0.2541& 0.3562& 0.9283&0.8951& - & - & - \\ 
        \midrule
        w/o \blue{FAE} & 0.2675& 0.3614& 0.9255&0.9134&v.s. Base Model & 59.7 v.s. \blue{40.3} & 76.4 v.s. \blue{23.6} \\ 
        w/o RefAdapter & 0.2640& 0.3856& 0.9217&0.9021&v.s. Base Model & 68.6 v.s. 31.4 & 52.2 v.s. 47.8 \\ 
        \midrule
        \textbf{Ours}& \textbf{0.2732}& \textbf{0.4103}& \textbf{0.9342}& \textbf{0.9162} &v.s. Base Model & \textbf{79.5 v.s. 20.5}& \textbf{78.3 v.s. 21.7}\\ 
        \bottomrule[1.5pt]
    \end{tabular}
    }
    \caption{Quantitative comparisons and human evaluations. We train an I2V version of Motion Director (MD-I2V) based on CogVideoX. The Base Model trains a set of learnable embeddings without incorporating both RefAdapter and FAE. ``$p_1$ v.s. $p_2$" means $p_1\%$ results of the first method are preferred.}
    \vspace{-20pt}
    \label{tab:comparison}
\end{table*}

\subsection{Implementation Details}
\noindent\textbf{Evaluation Dataset.}
We conduct experiments on a evaluation dataset of 250 video-image pairs, featuring 25 distinct action categories. Each action is transferred to 10 different target images, covering a wide range of human motions (e.g., yoga, fitness exercises) and animal motions (e.g., jumping, running). The target images include real humans, animals, animated, and game characters. This diversity ensures our dataset encompasses a broad spectrum of scenarios, allowing for a comprehensive evaluation of our method's generalization capabilities.

\noindent\textbf{Comparison Methods.}
Existing methods for action transfer include those based on predefined signals and global motion. Predefined signal methods are ineffective for non-human entities or subjects with significant skeletal differences, as shown in Figure \ref{fig:othermethods}. 
Therefore, we use the recent global motion transfer method, MotionDirector~\cite{tu2024motioneditor}, as our baseline. For a fair comparison, we reimplement MotionDirector on the stronger CogVideoX-I2V backbone (referred to as MD-I2V) with identical training settings to our methods. Additionally, we implement a base model that learns actions directly through standard learnable action embeddings, without using RefAdapter and \stagetwo (referred to as BaseModel).

\noindent\textbf{Training Details.}
For RefAdapter's training, we conduct a one-time 40,000-step training on Miradata \cite{ju2024miradata} with a learning rate of 1e-5 and a batch size of 8 with the AdamW optimizer. RefAdapter introduces 66M parameters, constituting 5\% of CogVideoX-I2V's total parameters.
Frequency-aware embeddings require 1,500 to 3,000 training steps on each reference video, depending on action complexity. In comparison, the Motion Director needs 3,000 steps for temporal LoRA and 300 for spatial LoRA.

\subsection{Quantitative Evaluation}

\label{subsec:quantative}

\noindent\textbf{Automatic Evaluations.} Following \cite{wang2024motion,jeong2023vmc,yatim2023space}, we employ \textit{Text Similarity}, \textit{Motion Fidelity}, and \textit{Temporal Consistency} to evaluate the semantic accuracy of the generated videos, the degree of motion alignment with the reference videos, and the temporal coherence, respectively. Furthermore, we introduce \textit{Appearance Consistency} to assess the consistency in appearance between the generated videos and the target image. Below, we provide a brief overview of these metrics.

\textit{Text Similarity.} It is calculated with CLIP~\cite{radford2021learning} frame-to-text similarity, reflecting the semantic alignment degree between the output video and the prompt.

\textit{Motion Fidelity.}
Introduced by~\cite{yatim2023space}, it utilizes tracklets computed by a tracking model~\cite{karaev2023cotracker}, measuring the similarity between the motion trajectories in unaligned videos.

\textit{Temporal Consistency.}
It measures the smoothness and coherence of a video sequence~\cite{jeong2023vmc, zhao2023motiondirector, wu2023tune, chen2023control}, quantified by the average similarity between the CLIP image features of all frame pairs within the output video.

\textit{Appearance Consistency.}
It reflects the appearance consistency between the output video and the target image, calculated as the average CLIP similarity between the first frame and the remaining frames of the output video.

\noindent\textbf{Human Evaluation.}
Following~\cite{zhao2023motiondirector}, we conduct a human evaluation with 5 raters who assessed each generated video for appearance consistency with the target image and motion consistency with the reference video. Each participant compares 50 randomly selected video pairs, each containing one video generated by a random method and one by the Base Model. Following~\cite{zhao2023motiondirector}, all methods are compared against the Base Model, serving as a solid baseline due to its comparable performance to MD-I2V.
In Table \ref{tab:comparison}, ``$p_1$ v.s. $p_2$" indicates that $p_1\%$ of the first method’s results are prefer over $p_2\%$ of the second method’s results.

\begin{figure*}
    \centering
    \includegraphics[width=\textwidth]{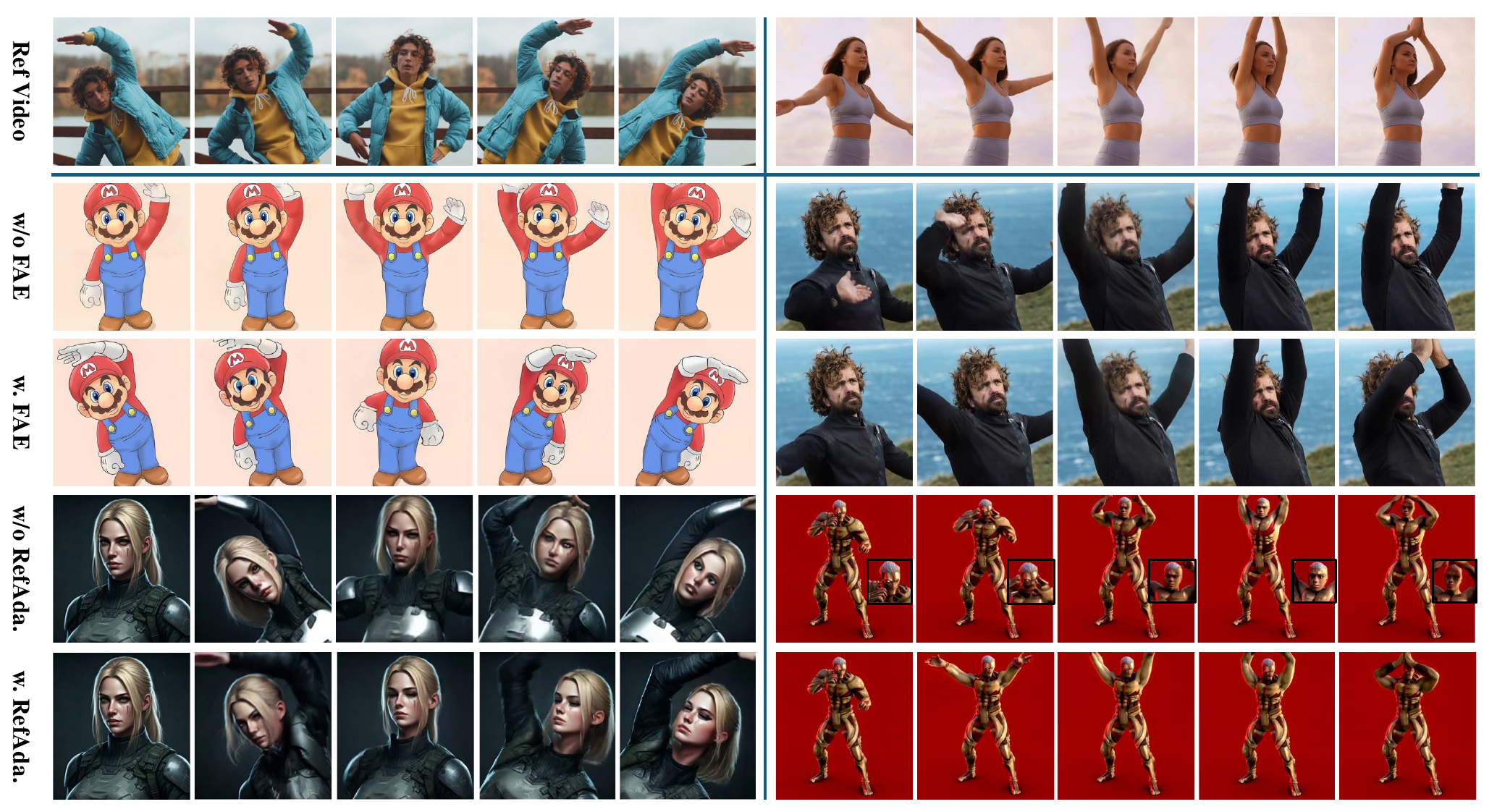}
    \vspace{-23pt}
    \caption{\textbf{Qualitative results of ablation study.}We ablate Frequency-aware Action Extraction (FAE) and RefAdapter, comparing the action transfer results from reference videos (Ref Video) to different subjects. Ablating FAE reduces action accuracy, demonstrating its effectiveness in action extraction. Ablating RefAdapter degrades both appearance consistency and action precision, proving its capability in spatial structure adaptation for cross-subject action transfer.}
    \label{fig:ablation}
    \vspace{-12pt}
\end{figure*}

\noindent\textbf{Results.} As shown in Table \ref{tab:comparison}, our method significantly outperforms baseline approaches in both motion fidelity and appearance consistency. This underscores the challenges of action transfer in heterogeneous scenarios and demonstrates our approach's effectiveness in balancing action accuracy with appearance consistency. 
Notably, our Motion Fidelity scores are generally lower than those in global motion tasks, as they are affected by layout consistency, whereas global motion tasks involve transferring motion to scenarios with identical layouts, making them not directly comparable to action transfer in heterogeneous scenarios.

\subsection{Qualitative Evaluation}
\label{subsec:qualitative}
Figure \ref{fig:comparison} shows a qualitative comparison with the baseline method.
MD-I2V struggles to replicate the reference video's motion accurately. In the first example, the man fails to stand up after squatting, and his arm movements do not match the original. In the second, he does not lift his leg as in the reference, and one eye closes in later frames. The Base Model also suffers from motion accuracy and appearance consistency issues.
In the first example, the man puts on clothes, deviating from the original image, and his final motion differs from the reference. In the second, his leg lift is exaggerated, and clothing-like folds appear in the final frames. In contrast, our method excels in both motion accuracy and appearance consistency.

\subsection{Ablation Study}

Table \ref{tab:comparison} and Figure \ref{fig:ablation} present our Ablation Study results. Quantitative data show that removing \stagetwo significantly decreases Motion Fidelity, highlighting its role in enhancing motion generation quality. This is corroborated by qualitative results, where two distinct actions transferred to different characters exhibit inconsistencies without \stagetwo. For example, in a stretching motion, the character merely raises their hand without proper bending or stretching, deviating from the reference video. Similar mismatches in the second example further emphasize \stagetwo's importance for motion consistency.

We also examined the impact of removing RefAdapter. Quantitative results indicate noticeable declines in both appearance consistency and motion fidelity, as RefAdapter ensures adaptability to varying spatial structures. Without it, the model struggles to adapt motion to target images with different spatial layouts, weakening appearance consistency. Qualitative results in Figure \ref{fig:ablation} support this: in the first example, discrepancies in the character's face and clothing are resolved with RefAdapter. In the second example, without RefAdapter, the output video fails to extend arms fully, maintaining them bent, and shows noticeable differences in facial details, underscoring RefAdapter's role in maintaining both motion and appearance consistency.

\section{discussion}
\label{sec:conclusion}
In this paper, we tackle the action transfer in heterogeneous scenarios, where the main difficulty is achieving precise action transfer for subjects with different spatial structures while maintaining appearance consistency. We introduce FlexiAct, a flexible and versatile approach that surpasses existing methods. Our RefAdapter adapts to various spatial structures and ensures appearance consistency, while Frequency-aware Action Extraction allows for precisely extracting action during the denoising process. Extensive experiments show that FlexiAct effectively balances action accuracy and appearance consistency across diverse spatial structures and domains.

Despite achieving precise action and appearance consistency, like~\cite{zhao2023motiondirector,wang2024motion,yatim2023space,jeong2023vmc}, our method requires optimization for each reference video. Developing feed-forward motion transfer methods for heterogeneous scenarios is a key direction for future work.

\paragraph{Acknowledgments.} This work was supported by Guangdong Natural Science Funds for Distinguished Young Scholar (No. 2025B1515020012).

\appendix
\begin{figure*}
    \centering
    \includegraphics[width=\textwidth]{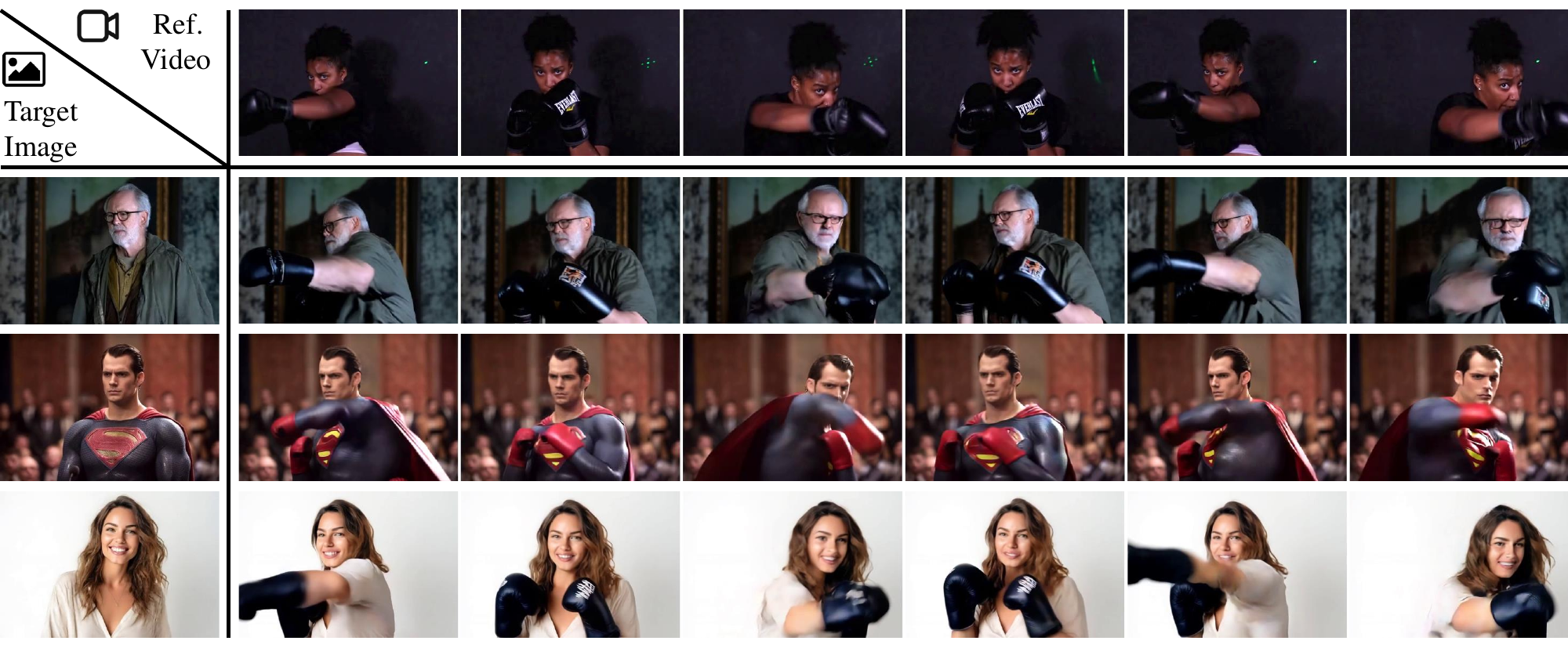}
    \caption{FlexiAct can transfer actions to diverse subjects while maintaining both appearance consistency with the target subject and action consistency with the reference video.}
    \label{fig:diff_h}
\end{figure*}
\begin{figure*}
    \centering
    \includegraphics[width=\textwidth]{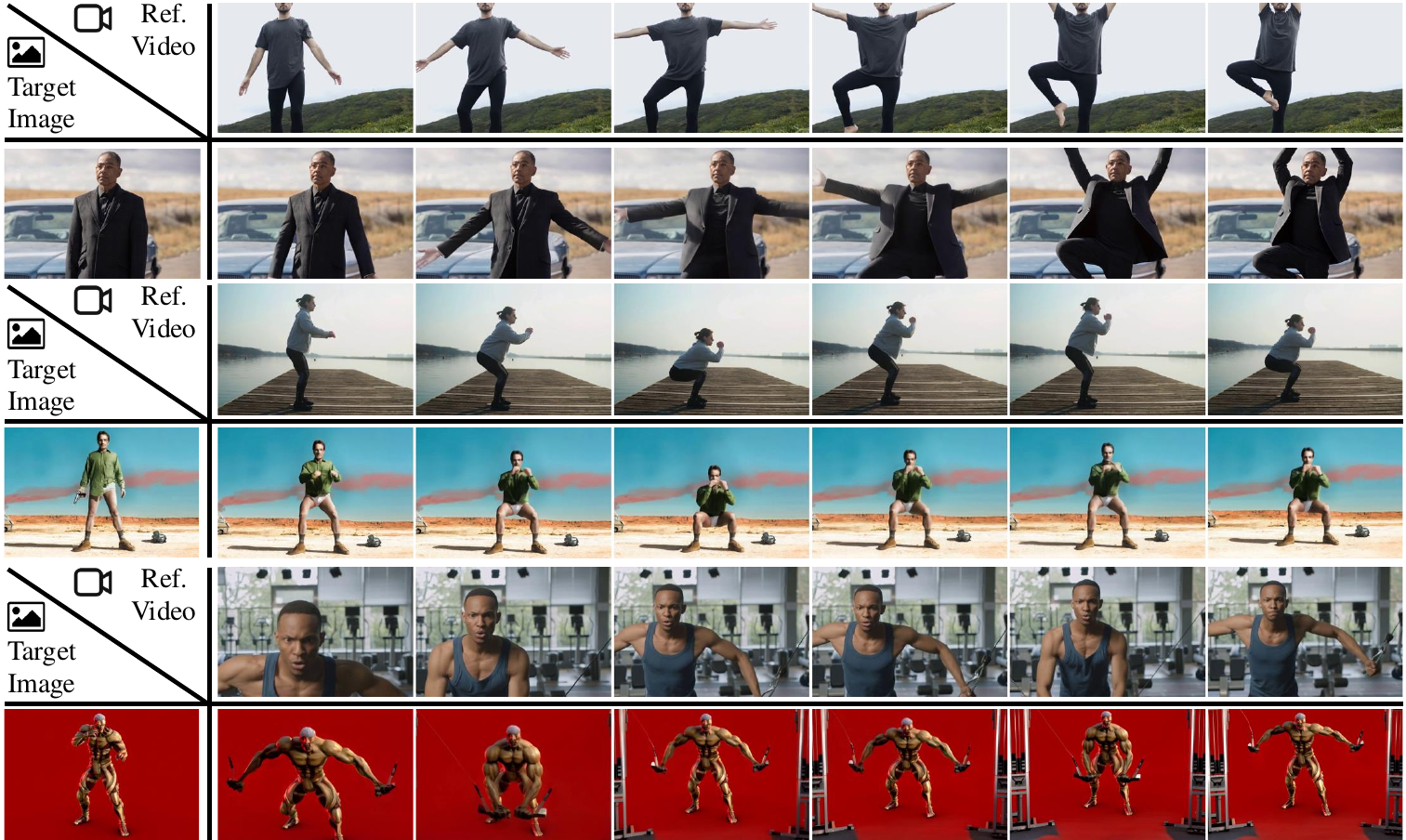}
    \caption{Examples of human action transfer using FlexiAct.}
    \label{fig:human}
\end{figure*}
\begin{figure*}
    \centering
    \includegraphics[width=\textwidth]{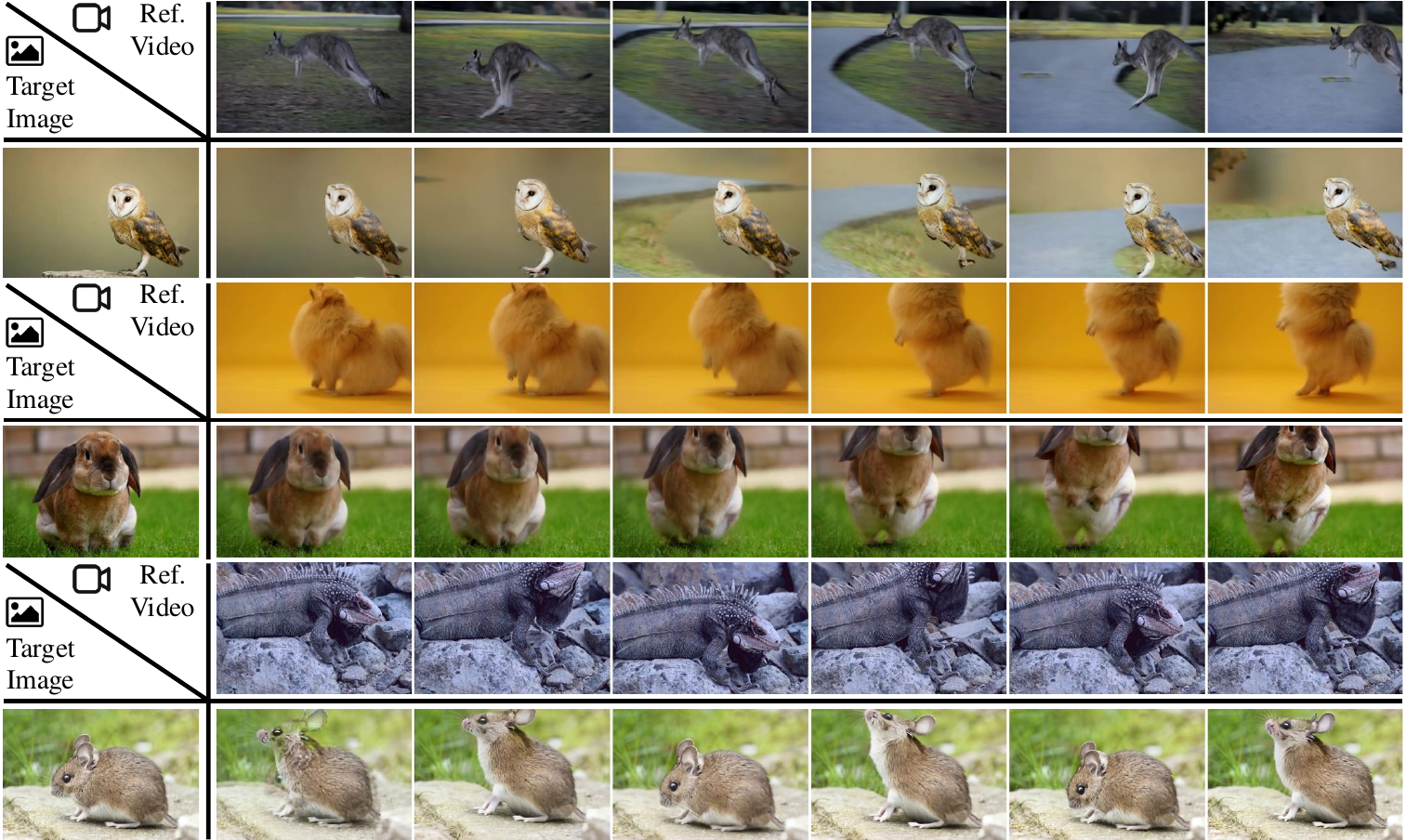}
    \vspace{-20pt}
    \caption{Examples of action transfer between animals using FlexiAct.}
    \label{fig:animal}
\end{figure*}
\begin{figure*}
    \centering
    \vspace{-8pt}
    \includegraphics[width=\textwidth]{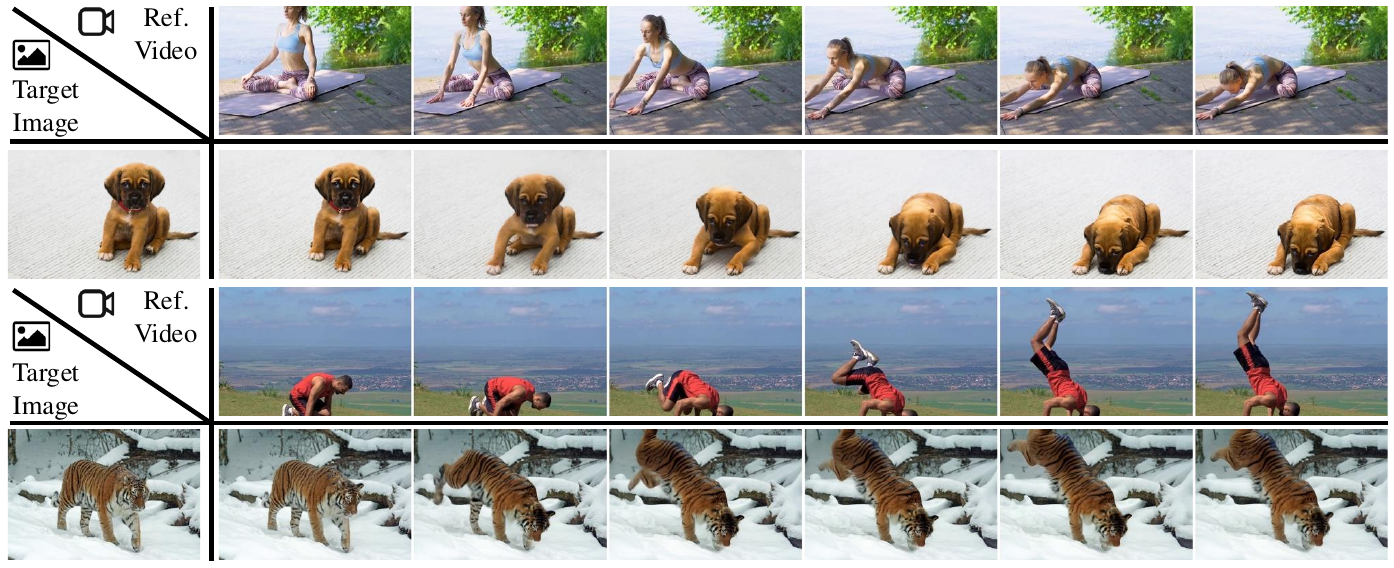}
    \vspace{-20pt}
    \caption{Examples of action transfer from humans to animals using FlexiAct.}
    \label{fig:h2animal}
\end{figure*}
\begin{figure*}
    \centering
    \vspace{-6pt}
    \includegraphics[width=\textwidth]{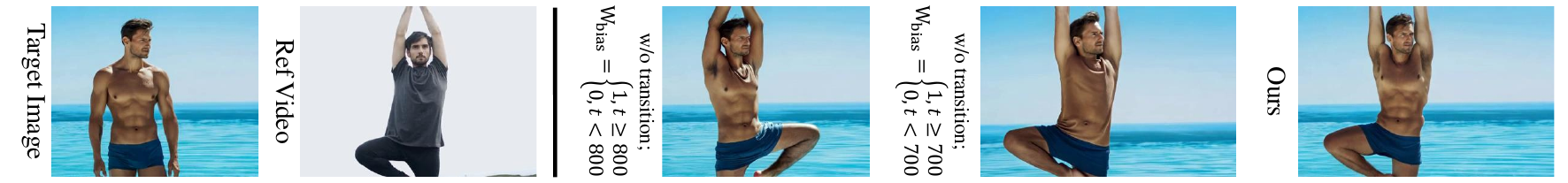}
    \vspace{-20pt}
    \caption{Ablation of bias transition.}
    \label{fig:supp_ablation}
\end{figure*}

\bibliographystyle{ACM-Reference-Format}
\bibliography{sample-bibliography}


\begin{thebibliography}{55}


\ifx \showCODEN    \undefined \def \showCODEN     #1{\unskip}     \fi
\ifx \showDOI      \undefined \def \showDOI       #1{#1}\fi
\ifx \showISBNx    \undefined \def \showISBNx     #1{\unskip}     \fi
\ifx \showISBNxiii \undefined \def \showISBNxiii  #1{\unskip}     \fi
\ifx \showISSN     \undefined \def \showISSN      #1{\unskip}     \fi
\ifx \showLCCN     \undefined \def \showLCCN      #1{\unskip}     \fi
\ifx \shownote     \undefined \def \shownote      #1{#1}          \fi
\ifx \showarticletitle \undefined \def \showarticletitle #1{#1}   \fi
\ifx \showURL      \undefined \def \showURL       {\relax}        \fi
\providecommand\bibfield[2]{#2}
\providecommand\bibinfo[2]{#2}
\providecommand\natexlab[1]{#1}
\providecommand\showeprint[2][]{arXiv:#2}

\bibitem[Bar-Tal et~al\mbox{.}(2024)]%
        {bar2024lumiere}
\bibfield{author}{\bibinfo{person}{Omer Bar-Tal}, \bibinfo{person}{Hila Chefer}, \bibinfo{person}{Omer Tov}, \bibinfo{person}{Charles Herrmann}, \bibinfo{person}{Roni Paiss}, \bibinfo{person}{Shiran Zada}, \bibinfo{person}{Ariel Ephrat}, \bibinfo{person}{Junhwa Hur}, \bibinfo{person}{Guanghui Liu}, \bibinfo{person}{Amit Raj}, {et~al\mbox{.}}} \bibinfo{year}{2024}\natexlab{}.
\newblock \showarticletitle{Lumiere: A space-time diffusion model for video generation}.
\newblock \bibinfo{journal}{\emph{arXiv preprint arXiv:2401.12945}} (\bibinfo{year}{2024}).
\newblock


\bibitem[Blattmann et~al\mbox{.}(2023)]%
        {blattmann2023align}
\bibfield{author}{\bibinfo{person}{Andreas Blattmann}, \bibinfo{person}{Robin Rombach}, \bibinfo{person}{Huan Ling}, \bibinfo{person}{Tim Dockhorn}, \bibinfo{person}{Seung~Wook Kim}, \bibinfo{person}{Sanja Fidler}, {and} \bibinfo{person}{Karsten Kreis}.} \bibinfo{year}{2023}\natexlab{}.
\newblock \showarticletitle{Align your Latents: High-Resolution Video Synthesis with Latent Diffusion Models}. In \bibinfo{booktitle}{\emph{Proc. CVPR}}.
\newblock


\bibitem[Brooks et~al\mbox{.}(2024)]%
        {videoworldsimulators2024}
\bibfield{author}{\bibinfo{person}{Tim Brooks}, \bibinfo{person}{Bill Peebles}, \bibinfo{person}{Connor Holmes}, \bibinfo{person}{Will DePue}, \bibinfo{person}{Yufei Guo}, \bibinfo{person}{Li Jing}, \bibinfo{person}{David Schnurr}, \bibinfo{person}{Joe Taylor}, \bibinfo{person}{Troy Luhman}, \bibinfo{person}{Eric Luhman}, \bibinfo{person}{Clarence Ng}, \bibinfo{person}{Ricky Wang}, {and} \bibinfo{person}{Aditya Ramesh}.} \bibinfo{year}{2024}\natexlab{}.
\newblock \showarticletitle{Video generation models as world simulators}.
\newblock  (\bibinfo{year}{2024}).
\newblock
\urldef\tempurl%
\url{https://openai.com/research/video-generation-models-as-world-simulators}
\showURL{%
\tempurl}


\bibitem[Bruce et~al\mbox{.}(2024)]%
        {bruce2024genie}
\bibfield{author}{\bibinfo{person}{Jake Bruce}, \bibinfo{person}{Michael~D Dennis}, \bibinfo{person}{Ashley Edwards}, \bibinfo{person}{Jack Parker-Holder}, \bibinfo{person}{Yuge Shi}, \bibinfo{person}{Edward Hughes}, \bibinfo{person}{Matthew Lai}, \bibinfo{person}{Aditi Mavalankar}, \bibinfo{person}{Richie Steigerwald}, \bibinfo{person}{Chris Apps}, {et~al\mbox{.}}} \bibinfo{year}{2024}\natexlab{}.
\newblock \showarticletitle{Genie: Generative interactive environments}. In \bibinfo{booktitle}{\emph{Forty-first International Conference on Machine Learning}}.
\newblock


\bibitem[Chen et~al\mbox{.}(2023b)]%
        {chen2023videocrafter1}
\bibfield{author}{\bibinfo{person}{Haoxin Chen}, \bibinfo{person}{Menghan Xia}, \bibinfo{person}{Yingqing He}, \bibinfo{person}{Yong Zhang}, \bibinfo{person}{Xiaodong Cun}, \bibinfo{person}{Shaoshu Yang}, \bibinfo{person}{Jinbo Xing}, \bibinfo{person}{Yaofang Liu}, \bibinfo{person}{Qifeng Chen}, \bibinfo{person}{Xintao Wang}, {et~al\mbox{.}}} \bibinfo{year}{2023}\natexlab{b}.
\newblock \showarticletitle{VideoCrafter1: Open Diffusion Models for High-Quality Video Generation}.
\newblock \bibinfo{journal}{\emph{arXiv preprint arXiv:2310.19512}} (\bibinfo{year}{2023}).
\newblock


\bibitem[Chen et~al\mbox{.}(2024)]%
        {chen2024videocrafter2}
\bibfield{author}{\bibinfo{person}{Haoxin Chen}, \bibinfo{person}{Yong Zhang}, \bibinfo{person}{Xiaodong Cun}, \bibinfo{person}{Menghan Xia}, \bibinfo{person}{Xintao Wang}, \bibinfo{person}{Chao Weng}, {and} \bibinfo{person}{Ying Shan}.} \bibinfo{year}{2024}\natexlab{}.
\newblock \showarticletitle{Videocrafter2: Overcoming data limitations for high-quality video diffusion models}. In \bibinfo{booktitle}{\emph{Proc. CVPR}}. \bibinfo{pages}{7310--7320}.
\newblock


\bibitem[Chen et~al\mbox{.}(2023a)]%
        {chen2023control}
\bibfield{author}{\bibinfo{person}{Weifeng Chen}, \bibinfo{person}{Jie Wu}, \bibinfo{person}{Pan Xie}, \bibinfo{person}{Hefeng Wu}, \bibinfo{person}{Jiashi Li}, \bibinfo{person}{Xin Xia}, \bibinfo{person}{Xuefeng Xiao}, {and} \bibinfo{person}{Liang Lin}.} \bibinfo{year}{2023}\natexlab{a}.
\newblock \showarticletitle{Control-A-Video: Controllable Text-to-Video Generation with Diffusion Models}.
\newblock \bibinfo{journal}{\emph{arXiv preprint arXiv:2305.13840}} (\bibinfo{year}{2023}).
\newblock


\bibitem[Dao and Gu(2024)]%
        {mamba2}
\bibfield{author}{\bibinfo{person}{Tri Dao} {and} \bibinfo{person}{Albert Gu}.} \bibinfo{year}{2024}\natexlab{}.
\newblock \showarticletitle{Transformers are {SSM}s: Generalized Models and Efficient Algorithms Through Structured State Space Duality}. In \bibinfo{booktitle}{\emph{ICML}}.
\newblock


\bibitem[Esser et~al\mbox{.}(2023)]%
        {esser2023structure}
\bibfield{author}{\bibinfo{person}{Patrick Esser}, \bibinfo{person}{Johnathan Chiu}, \bibinfo{person}{Parmida Atighehchian}, \bibinfo{person}{Jonathan Granskog}, {and} \bibinfo{person}{Anastasis Germanidis}.} \bibinfo{year}{2023}\natexlab{}.
\newblock \showarticletitle{Structure and content-guided video synthesis with diffusion models}. In \bibinfo{booktitle}{\emph{Proc. ICCV}}. \bibinfo{pages}{7346--7356}.
\newblock


\bibitem[Esser et~al\mbox{.}(2024)]%
        {esser2024scaling}
\bibfield{author}{\bibinfo{person}{Patrick Esser}, \bibinfo{person}{Sumith Kulal}, \bibinfo{person}{Andreas Blattmann}, \bibinfo{person}{Rahim Entezari}, \bibinfo{person}{Jonas M{\"u}ller}, \bibinfo{person}{Harry Saini}, \bibinfo{person}{Yam Levi}, \bibinfo{person}{Dominik Lorenz}, \bibinfo{person}{Axel Sauer}, \bibinfo{person}{Frederic Boesel}, {et~al\mbox{.}}} \bibinfo{year}{2024}\natexlab{}.
\newblock \showarticletitle{Scaling rectified flow transformers for high-resolution image synthesis}. In \bibinfo{booktitle}{\emph{Forty-first international conference on machine learning}}.
\newblock


\bibitem[Gal et~al\mbox{.}(2022)]%
        {gal2022textual}
\bibfield{author}{\bibinfo{person}{Rinon Gal}, \bibinfo{person}{Yuval Alaluf}, \bibinfo{person}{Yuval Atzmon}, \bibinfo{person}{Or Patashnik}, \bibinfo{person}{Amit~H Bermano}, \bibinfo{person}{Gal Chechik}, {and} \bibinfo{person}{Daniel Cohen-Or}.} \bibinfo{year}{2022}\natexlab{}.
\newblock \showarticletitle{An image is worth one word: Personalizing text-to-image generation using textual inversion}.
\newblock \bibinfo{journal}{\emph{arXiv preprint arXiv:2208.01618}} (\bibinfo{year}{2022}).
\newblock


\bibitem[Goodfellow et~al\mbox{.}(2020)]%
        {goodfellow2020generative}
\bibfield{author}{\bibinfo{person}{Ian Goodfellow}, \bibinfo{person}{Jean Pouget-Abadie}, \bibinfo{person}{Mehdi Mirza}, \bibinfo{person}{Bing Xu}, \bibinfo{person}{David Warde-Farley}, \bibinfo{person}{Sherjil Ozair}, \bibinfo{person}{Aaron Courville}, {and} \bibinfo{person}{Yoshua Bengio}.} \bibinfo{year}{2020}\natexlab{}.
\newblock \showarticletitle{Generative adversarial networks}.
\newblock \bibinfo{journal}{\emph{Commun. ACM}} (\bibinfo{year}{2020}).
\newblock


\bibitem[Guo et~al\mbox{.}(2023)]%
        {guo2023animatediff}
\bibfield{author}{\bibinfo{person}{Yuwei Guo}, \bibinfo{person}{Ceyuan Yang}, \bibinfo{person}{Anyi Rao}, \bibinfo{person}{Yaohui Wang}, \bibinfo{person}{Yu Qiao}, \bibinfo{person}{Dahua Lin}, {and} \bibinfo{person}{Bo Dai}.} \bibinfo{year}{2023}\natexlab{}.
\newblock \showarticletitle{Animatediff: Animate your personalized text-to-image diffusion models without specific tuning}.
\newblock \bibinfo{journal}{\emph{arXiv preprint arXiv:2307.04725}} (\bibinfo{year}{2023}).
\newblock


\bibitem[He et~al\mbox{.}(2022)]%
        {he2022lvdm}
\bibfield{author}{\bibinfo{person}{Yingqing He}, \bibinfo{person}{Tianyu Yang}, \bibinfo{person}{Yong Zhang}, \bibinfo{person}{Ying Shan}, {and} \bibinfo{person}{Qifeng Chen}.} \bibinfo{year}{2022}\natexlab{}.
\newblock \showarticletitle{Latent Video Diffusion Models for High-Fidelity Long Video Generation}.
\newblock  (\bibinfo{year}{2022}).
\newblock
\showeprint[arxiv]{2211.13221}~[cs.CV]


\bibitem[Ho et~al\mbox{.}(2020)]%
        {ho2020denoising}
\bibfield{author}{\bibinfo{person}{Jonathan Ho}, \bibinfo{person}{Ajay Jain}, {and} \bibinfo{person}{Pieter Abbeel}.} \bibinfo{year}{2020}\natexlab{}.
\newblock \showarticletitle{Denoising diffusion probabilistic models}.
\newblock \bibinfo{journal}{\emph{Proc. NeurIPS}}  \bibinfo{volume}{33} (\bibinfo{year}{2020}), \bibinfo{pages}{6840--6851}.
\newblock


\bibitem[Hu et~al\mbox{.}(2021)]%
        {hu2021lora}
\bibfield{author}{\bibinfo{person}{Edward~J Hu}, \bibinfo{person}{Yelong Shen}, \bibinfo{person}{Phillip Wallis}, \bibinfo{person}{Zeyuan Allen-Zhu}, \bibinfo{person}{Yuanzhi Li}, \bibinfo{person}{Shean Wang}, \bibinfo{person}{Lu Wang}, {and} \bibinfo{person}{Weizhu Chen}.} \bibinfo{year}{2021}\natexlab{}.
\newblock \showarticletitle{Lora: Low-rank adaptation of large language models}.
\newblock \bibinfo{journal}{\emph{arXiv preprint arXiv:2106.09685}} (\bibinfo{year}{2021}).
\newblock


\bibitem[Hu(2024)]%
        {hu2024animate}
\bibfield{author}{\bibinfo{person}{Li Hu}.} \bibinfo{year}{2024}\natexlab{}.
\newblock \showarticletitle{Animate anyone: Consistent and controllable image-to-video synthesis for character animation}. In \bibinfo{booktitle}{\emph{Proc. CVPR}}. \bibinfo{pages}{8153--8163}.
\newblock


\bibitem[Huang et~al\mbox{.}(2021)]%
        {huang2021few}
\bibfield{author}{\bibinfo{person}{Zhichao Huang}, \bibinfo{person}{Xintong Han}, \bibinfo{person}{Jia Xu}, {and} \bibinfo{person}{Tong Zhang}.} \bibinfo{year}{2021}\natexlab{}.
\newblock \showarticletitle{Few-shot human motion transfer by personalized geometry and texture modeling}. In \bibinfo{booktitle}{\emph{CVPR}}.
\newblock


\bibitem[Jeong et~al\mbox{.}(2024)]%
        {jeong2024dreammotion}
\bibfield{author}{\bibinfo{person}{Hyeonho Jeong}, \bibinfo{person}{Jinho Chang}, \bibinfo{person}{Geon~Yeong Park}, {and} \bibinfo{person}{Jong~Chul Ye}.} \bibinfo{year}{2024}\natexlab{}.
\newblock \showarticletitle{DreamMotion: Space-Time Self-Similarity Score Distillation for Zero-Shot Video Editing}.
\newblock \bibinfo{journal}{\emph{arXiv preprint arXiv:2403.12002}} (\bibinfo{year}{2024}).
\newblock


\bibitem[Jeong et~al\mbox{.}(2023)]%
        {jeong2023vmc}
\bibfield{author}{\bibinfo{person}{Hyeonho Jeong}, \bibinfo{person}{Geon~Yeong Park}, {and} \bibinfo{person}{Jong~Chul Ye}.} \bibinfo{year}{2023}\natexlab{}.
\newblock \showarticletitle{VMC: Video Motion Customization using Temporal Attention Adaption for Text-to-Video Diffusion Models}.
\newblock \bibinfo{journal}{\emph{arXiv preprint arXiv:2312.00845}} (\bibinfo{year}{2023}).
\newblock


\bibitem[Ju et~al\mbox{.}(2024)]%
        {ju2024miradata}
\bibfield{author}{\bibinfo{person}{Xuan Ju}, \bibinfo{person}{Yiming Gao}, \bibinfo{person}{Zhaoyang Zhang}, \bibinfo{person}{Ziyang Yuan}, \bibinfo{person}{Xintao Wang}, \bibinfo{person}{Ailing Zeng}, \bibinfo{person}{Yu Xiong}, \bibinfo{person}{Qiang Xu}, {and} \bibinfo{person}{Ying Shan}.} \bibinfo{year}{2024}\natexlab{}.
\newblock \showarticletitle{Miradata: A large-scale video dataset with long durations and structured captions}.
\newblock \bibinfo{journal}{\emph{arXiv preprint arXiv:2407.06358}} (\bibinfo{year}{2024}).
\newblock


\bibitem[Karaev et~al\mbox{.}(2023)]%
        {karaev2023cotracker}
\bibfield{author}{\bibinfo{person}{Nikita Karaev}, \bibinfo{person}{Ignacio Rocco}, \bibinfo{person}{Benjamin Graham}, \bibinfo{person}{Natalia Neverova}, \bibinfo{person}{Andrea Vedaldi}, {and} \bibinfo{person}{Christian Rupprecht}.} \bibinfo{year}{2023}\natexlab{}.
\newblock \showarticletitle{Cotracker: It is better to track together}.
\newblock \bibinfo{journal}{\emph{arXiv preprint arXiv:2307.07635}} (\bibinfo{year}{2023}).
\newblock


\bibitem[Lin et~al\mbox{.}(2024)]%
        {lin2024common}
\bibfield{author}{\bibinfo{person}{Shanchuan Lin}, \bibinfo{person}{Bingchen Liu}, \bibinfo{person}{Jiashi Li}, {and} \bibinfo{person}{Xiao Yang}.} \bibinfo{year}{2024}\natexlab{}.
\newblock \showarticletitle{Common diffusion noise schedules and sample steps are flawed}. In \bibinfo{booktitle}{\emph{Proceedings of the IEEE/CVF winter conference on applications of computer vision}}. \bibinfo{pages}{5404--5411}.
\newblock


\bibitem[Ling et~al\mbox{.}(2024)]%
        {ling2024motionclone}
\bibfield{author}{\bibinfo{person}{Pengyang Ling}, \bibinfo{person}{Jiazi Bu}, \bibinfo{person}{Pan Zhang}, \bibinfo{person}{Xiaoyi Dong}, \bibinfo{person}{Yuhang Zang}, \bibinfo{person}{Tong Wu}, \bibinfo{person}{Huaian Chen}, \bibinfo{person}{Jiaqi Wang}, {and} \bibinfo{person}{Yi Jin}.} \bibinfo{year}{2024}\natexlab{}.
\newblock \showarticletitle{MotionClone: Training-Free Motion Cloning for Controllable Video Generation}.
\newblock \bibinfo{journal}{\emph{arXiv preprint arXiv:2406.05338}} (\bibinfo{year}{2024}).
\newblock


\bibitem[Lu et~al\mbox{.}(2024)]%
        {lu2024freelong}
\bibfield{author}{\bibinfo{person}{Yu Lu}, \bibinfo{person}{Yuanzhi Liang}, \bibinfo{person}{Linchao Zhu}, {and} \bibinfo{person}{Yi Yang}.} \bibinfo{year}{2024}\natexlab{}.
\newblock \showarticletitle{Freelong: Training-free long video generation with spectralblend temporal attention}.
\newblock \bibinfo{journal}{\emph{arXiv preprint arXiv:2407.19918}} (\bibinfo{year}{2024}).
\newblock


\bibitem[Ma et~al\mbox{.}(2024)]%
        {ma2024latte}
\bibfield{author}{\bibinfo{person}{Xin Ma}, \bibinfo{person}{Yaohui Wang}, \bibinfo{person}{Gengyun Jia}, \bibinfo{person}{Xinyuan Chen}, \bibinfo{person}{Ziwei Liu}, \bibinfo{person}{Yuan-Fang Li}, \bibinfo{person}{Cunjian Chen}, {and} \bibinfo{person}{Yu Qiao}.} \bibinfo{year}{2024}\natexlab{}.
\newblock \showarticletitle{Latte: Latent diffusion transformer for video generation}.
\newblock \bibinfo{journal}{\emph{arXiv preprint arXiv:2401.03048}} (\bibinfo{year}{2024}).
\newblock


\bibitem[Peng et~al\mbox{.}(2024)]%
        {peng2024controlnext}
\bibfield{author}{\bibinfo{person}{Bohao Peng}, \bibinfo{person}{Jian Wang}, \bibinfo{person}{Yuechen Zhang}, \bibinfo{person}{Wenbo Li}, \bibinfo{person}{Ming-Chang Yang}, {and} \bibinfo{person}{Jiaya Jia}.} \bibinfo{year}{2024}\natexlab{}.
\newblock \showarticletitle{ControlNeXt: Powerful and Efficient Control for Image and Video Generation}.
\newblock \bibinfo{journal}{\emph{arXiv preprint arXiv:2408.06070}} (\bibinfo{year}{2024}).
\newblock


\bibitem[Qiu et~al\mbox{.}(2023)]%
        {qiu2023freenoise}
\bibfield{author}{\bibinfo{person}{Haonan Qiu}, \bibinfo{person}{Menghan Xia}, \bibinfo{person}{Yong Zhang}, \bibinfo{person}{Yingqing He}, \bibinfo{person}{Xintao Wang}, \bibinfo{person}{Ying Shan}, {and} \bibinfo{person}{Ziwei Liu}.} \bibinfo{year}{2023}\natexlab{}.
\newblock \showarticletitle{Freenoise: Tuning-free longer video diffusion via noise rescheduling}.
\newblock \bibinfo{journal}{\emph{arXiv preprint arXiv:2310.15169}} (\bibinfo{year}{2023}).
\newblock


\bibitem[Radford et~al\mbox{.}(2021)]%
        {radford2021learning}
\bibfield{author}{\bibinfo{person}{Alec Radford}, \bibinfo{person}{Jong~Wook Kim}, \bibinfo{person}{Chris Hallacy}, \bibinfo{person}{Aditya Ramesh}, \bibinfo{person}{Gabriel Goh}, \bibinfo{person}{Sandhini Agarwal}, \bibinfo{person}{Girish Sastry}, \bibinfo{person}{Amanda Askell}, \bibinfo{person}{Pamela Mishkin}, \bibinfo{person}{Jack Clark}, {et~al\mbox{.}}} \bibinfo{year}{2021}\natexlab{}.
\newblock \showarticletitle{Learning transferable visual models from natural language supervision}. In \bibinfo{booktitle}{\emph{Proc. ICML}}. PMLR, \bibinfo{pages}{8748--8763}.
\newblock


\bibitem[Si et~al\mbox{.}(2024)]%
        {si2024freeu}
\bibfield{author}{\bibinfo{person}{Chenyang Si}, \bibinfo{person}{Ziqi Huang}, \bibinfo{person}{Yuming Jiang}, {and} \bibinfo{person}{Ziwei Liu}.} \bibinfo{year}{2024}\natexlab{}.
\newblock \showarticletitle{Freeu: Free lunch in diffusion u-net}. In \bibinfo{booktitle}{\emph{Proceedings of the IEEE/CVF Conference on Computer Vision and Pattern Recognition}}. \bibinfo{pages}{4733--4743}.
\newblock


\bibitem[Siarohin et~al\mbox{.}(2019)]%
        {Siarohin_2019_NeurIPS}
\bibfield{author}{\bibinfo{person}{Aliaksandr Siarohin}, \bibinfo{person}{Stéphane Lathuilière}, \bibinfo{person}{Sergey Tulyakov}, \bibinfo{person}{Elisa Ricci}, {and} \bibinfo{person}{Nicu Sebe}.} \bibinfo{year}{2019}\natexlab{}.
\newblock \showarticletitle{First Order Motion Model for Image Animation}. In \bibinfo{booktitle}{\emph{NeurIPS}}.
\newblock


\bibitem[Siarohin et~al\mbox{.}(2021)]%
        {siarohin2021motion}
\bibfield{author}{\bibinfo{person}{Aliaksandr Siarohin}, \bibinfo{person}{Oliver~J Woodford}, \bibinfo{person}{Jian Ren}, \bibinfo{person}{Menglei Chai}, {and} \bibinfo{person}{Sergey Tulyakov}.} \bibinfo{year}{2021}\natexlab{}.
\newblock \showarticletitle{Motion representations for articulated animation}. In \bibinfo{booktitle}{\emph{CVPR}}.
\newblock


\bibitem[Tu et~al\mbox{.}(2024a)]%
        {tu2024motioneditor}
\bibfield{author}{\bibinfo{person}{Shuyuan Tu}, \bibinfo{person}{Qi Dai}, \bibinfo{person}{Zhi-Qi Cheng}, \bibinfo{person}{Han Hu}, \bibinfo{person}{Xintong Han}, \bibinfo{person}{Zuxuan Wu}, {and} \bibinfo{person}{Yu-Gang Jiang}.} \bibinfo{year}{2024}\natexlab{a}.
\newblock \showarticletitle{Motioneditor: Editing video motion via content-aware diffusion}. In \bibinfo{booktitle}{\emph{CVPR}}.
\newblock


\bibitem[Tu et~al\mbox{.}(2024b)]%
        {tu2024stableanimator}
\bibfield{author}{\bibinfo{person}{Shuyuan Tu}, \bibinfo{person}{Zhen Xing}, \bibinfo{person}{Xintong Han}, \bibinfo{person}{Zhi-Qi Cheng}, \bibinfo{person}{Qi Dai}, \bibinfo{person}{Chong Luo}, {and} \bibinfo{person}{Zuxuan Wu}.} \bibinfo{year}{2024}\natexlab{b}.
\newblock \showarticletitle{StableAnimator: High-Quality Identity-Preserving Human Image Animation}.
\newblock \bibinfo{journal}{\emph{arXiv preprint arXiv:2411.17697}} (\bibinfo{year}{2024}).
\newblock


\bibitem[Wang et~al\mbox{.}(2023c)]%
        {wang2023modelscope}
\bibfield{author}{\bibinfo{person}{Jiuniu Wang}, \bibinfo{person}{Hangjie Yuan}, \bibinfo{person}{Dayou Chen}, \bibinfo{person}{Yingya Zhang}, \bibinfo{person}{Xiang Wang}, {and} \bibinfo{person}{Shiwei Zhang}.} \bibinfo{year}{2023}\natexlab{c}.
\newblock \showarticletitle{Modelscope text-to-video technical report}.
\newblock \bibinfo{journal}{\emph{arXiv preprint arXiv:2308.06571}} (\bibinfo{year}{2023}).
\newblock


\bibitem[Wang et~al\mbox{.}(2024b)]%
        {wang2024motion}
\bibfield{author}{\bibinfo{person}{Luozhou Wang}, \bibinfo{person}{Ziyang Mai}, \bibinfo{person}{Guibao Shen}, \bibinfo{person}{Yixun Liang}, \bibinfo{person}{Xin Tao}, \bibinfo{person}{Pengfei Wan}, \bibinfo{person}{Di Zhang}, \bibinfo{person}{Yijun Li}, {and} \bibinfo{person}{Yingcong Chen}.} \bibinfo{year}{2024}\natexlab{b}.
\newblock \showarticletitle{Motion inversion for video customization}.
\newblock \bibinfo{journal}{\emph{arXiv preprint arXiv:2403.20193}} (\bibinfo{year}{2024}).
\newblock


\bibitem[Wang et~al\mbox{.}(2024a)]%
        {wang2024disco}
\bibfield{author}{\bibinfo{person}{Tan Wang}, \bibinfo{person}{Linjie Li}, \bibinfo{person}{Kevin Lin}, \bibinfo{person}{Yuanhao Zhai}, \bibinfo{person}{Chung-Ching Lin}, \bibinfo{person}{Zhengyuan Yang}, \bibinfo{person}{Hanwang Zhang}, \bibinfo{person}{Zicheng Liu}, {and} \bibinfo{person}{Lijuan Wang}.} \bibinfo{year}{2024}\natexlab{a}.
\newblock \showarticletitle{Disco: Disentangled control for realistic human dance generation}. In \bibinfo{booktitle}{\emph{CVPR}}.
\newblock


\bibitem[Wang et~al\mbox{.}(2023b)]%
        {wang2023videofactory}
\bibfield{author}{\bibinfo{person}{Wenjing Wang}, \bibinfo{person}{Huan Yang}, \bibinfo{person}{Zixi Tuo}, \bibinfo{person}{Huiguo He}, \bibinfo{person}{Junchen Zhu}, \bibinfo{person}{Jianlong Fu}, {and} \bibinfo{person}{Jiaying Liu}.} \bibinfo{year}{2023}\natexlab{b}.
\newblock \showarticletitle{VideoFactory: Swap Attention in Spatiotemporal Diffusions for Text-to-Video Generation}.
\newblock \bibinfo{journal}{\emph{arXiv preprint arXiv:2305.10874}} (\bibinfo{year}{2023}).
\newblock


\bibitem[Wang et~al\mbox{.}(2023d)]%
        {wang2023videocomposer}
\bibfield{author}{\bibinfo{person}{Xiang Wang}, \bibinfo{person}{Hangjie Yuan}, \bibinfo{person}{Shiwei Zhang}, \bibinfo{person}{Dayou Chen}, \bibinfo{person}{Jiuniu Wang}, \bibinfo{person}{Yingya Zhang}, \bibinfo{person}{Yujun Shen}, \bibinfo{person}{Deli Zhao}, {and} \bibinfo{person}{Jingren Zhou}.} \bibinfo{year}{2023}\natexlab{d}.
\newblock \showarticletitle{VideoComposer: Compositional Video Synthesis with Motion Controllability}.
\newblock \bibinfo{journal}{\emph{arXiv preprint arXiv:2306.02018}} (\bibinfo{year}{2023}).
\newblock


\bibitem[Wang et~al\mbox{.}(2024c)]%
        {wang2024unianimate}
\bibfield{author}{\bibinfo{person}{Xiang Wang}, \bibinfo{person}{Shiwei Zhang}, \bibinfo{person}{Changxin Gao}, \bibinfo{person}{Jiayu Wang}, \bibinfo{person}{Xiaoqiang Zhou}, \bibinfo{person}{Yingya Zhang}, \bibinfo{person}{Luxin Yan}, {and} \bibinfo{person}{Nong Sang}.} \bibinfo{year}{2024}\natexlab{c}.
\newblock \showarticletitle{UniAnimate: Taming Unified Video Diffusion Models for Consistent Human Image Animation}.
\newblock \bibinfo{journal}{\emph{arXiv preprint arXiv:2406.01188}} (\bibinfo{year}{2024}).
\newblock


\bibitem[Wang et~al\mbox{.}(2023a)]%
        {wang2023lavie}
\bibfield{author}{\bibinfo{person}{Yaohui Wang}, \bibinfo{person}{Xinyuan Chen}, \bibinfo{person}{Xin Ma}, \bibinfo{person}{Shangchen Zhou}, \bibinfo{person}{Ziqi Huang}, \bibinfo{person}{Yi Wang}, \bibinfo{person}{Ceyuan Yang}, \bibinfo{person}{Yinan He}, \bibinfo{person}{Jiashuo Yu}, \bibinfo{person}{Peiqing Yang}, {et~al\mbox{.}}} \bibinfo{year}{2023}\natexlab{a}.
\newblock \showarticletitle{LAVIE: High-Quality Video Generation with Cascaded Latent Diffusion Models}.
\newblock \bibinfo{journal}{\emph{arXiv preprint arXiv:2309.15103}} (\bibinfo{year}{2023}).
\newblock


\bibitem[Wu et~al\mbox{.}(2023a)]%
        {wu2023tune}
\bibfield{author}{\bibinfo{person}{Jay~Zhangjie Wu}, \bibinfo{person}{Yixiao Ge}, \bibinfo{person}{Xintao Wang}, \bibinfo{person}{Stan~Weixian Lei}, \bibinfo{person}{Yuchao Gu}, \bibinfo{person}{Yufei Shi}, \bibinfo{person}{Wynne Hsu}, \bibinfo{person}{Ying Shan}, \bibinfo{person}{Xiaohu Qie}, {and} \bibinfo{person}{Mike~Zheng Shou}.} \bibinfo{year}{2023}\natexlab{a}.
\newblock \showarticletitle{Tune-a-video: One-shot tuning of image diffusion models for text-to-video generation}. In \bibinfo{booktitle}{\emph{Proceedings of the IEEE/CVF International Conference on Computer Vision}}. \bibinfo{pages}{7623--7633}.
\newblock


\bibitem[Wu et~al\mbox{.}(2023b)]%
        {wu2023freeinit}
\bibfield{author}{\bibinfo{person}{Tianxing Wu}, \bibinfo{person}{Chenyang Si}, \bibinfo{person}{Yuming Jiang}, \bibinfo{person}{Ziqi Huang}, {and} \bibinfo{person}{Ziwei Liu}.} \bibinfo{year}{2023}\natexlab{b}.
\newblock \showarticletitle{Freeinit: Bridging initialization gap in video diffusion models}.
\newblock \bibinfo{journal}{\emph{arXiv preprint arXiv:2312.07537}} (\bibinfo{year}{2023}).
\newblock


\bibitem[Xu et~al\mbox{.}(2024)]%
        {xu2024magicanimate}
\bibfield{author}{\bibinfo{person}{Zhongcong Xu}, \bibinfo{person}{Jianfeng Zhang}, \bibinfo{person}{Jun~Hao Liew}, \bibinfo{person}{Hanshu Yan}, \bibinfo{person}{Jia-Wei Liu}, \bibinfo{person}{Chenxu Zhang}, \bibinfo{person}{Jiashi Feng}, {and} \bibinfo{person}{Mike~Zheng Shou}.} \bibinfo{year}{2024}\natexlab{}.
\newblock \showarticletitle{Magicanimate: Temporally consistent human image animation using diffusion model}. In \bibinfo{booktitle}{\emph{CVPR}}.
\newblock


\bibitem[Yang et~al\mbox{.}(2024)]%
        {yang2024cogvideox}
\bibfield{author}{\bibinfo{person}{Zhuoyi Yang}, \bibinfo{person}{Jiayan Teng}, \bibinfo{person}{Wendi Zheng}, \bibinfo{person}{Ming Ding}, \bibinfo{person}{Shiyu Huang}, \bibinfo{person}{Jiazheng Xu}, \bibinfo{person}{Yuanming Yang}, \bibinfo{person}{Wenyi Hong}, \bibinfo{person}{Xiaohan Zhang}, \bibinfo{person}{Guanyu Feng}, {et~al\mbox{.}}} \bibinfo{year}{2024}\natexlab{}.
\newblock \showarticletitle{CogVideoX: Text-to-Video Diffusion Models with An Expert Transformer}.
\newblock \bibinfo{journal}{\emph{arXiv preprint arXiv:2408.06072}} (\bibinfo{year}{2024}).
\newblock


\bibitem[Yatim et~al\mbox{.}(2023)]%
        {yatim2023space}
\bibfield{author}{\bibinfo{person}{Danah Yatim}, \bibinfo{person}{Rafail Fridman}, \bibinfo{person}{Omer~Bar Tal}, \bibinfo{person}{Yoni Kasten}, {and} \bibinfo{person}{Tali Dekel}.} \bibinfo{year}{2023}\natexlab{}.
\newblock \showarticletitle{Space-Time Diffusion Features for Zero-Shot Text-Driven Motion Transfer}.
\newblock \bibinfo{journal}{\emph{arXiv preprint arXiv:2311.17009}} (\bibinfo{year}{2023}).
\newblock


\bibitem[Ye et~al\mbox{.}(2023a)]%
        {ye2023ip-adapter}
\bibfield{author}{\bibinfo{person}{Hu Ye}, \bibinfo{person}{Jun Zhang}, \bibinfo{person}{Sibo Liu}, \bibinfo{person}{Xiao Han}, {and} \bibinfo{person}{Wei Yang}.} \bibinfo{year}{2023}\natexlab{a}.
\newblock \showarticletitle{IP-Adapter: Text Compatible Image Prompt Adapter for Text-to-Image Diffusion Models}.
\newblock \bibinfo{journal}{\emph{arXiv preprint arxiv:2308.06721}} (\bibinfo{year}{2023}).
\newblock


\bibitem[Ye et~al\mbox{.}(2023b)]%
        {ye2023ip}
\bibfield{author}{\bibinfo{person}{Hu Ye}, \bibinfo{person}{Jun Zhang}, \bibinfo{person}{Sibo Liu}, \bibinfo{person}{Xiao Han}, {and} \bibinfo{person}{Wei Yang}.} \bibinfo{year}{2023}\natexlab{b}.
\newblock \showarticletitle{Ip-adapter: Text compatible image prompt adapter for text-to-image diffusion models}.
\newblock \bibinfo{journal}{\emph{arXiv preprint arXiv:2308.06721}} (\bibinfo{year}{2023}).
\newblock


\bibitem[Yuan et~al\mbox{.}(2024)]%
        {yuan2024instructvideo}
\bibfield{author}{\bibinfo{person}{Hangjie Yuan}, \bibinfo{person}{Shiwei Zhang}, \bibinfo{person}{Xiang Wang}, \bibinfo{person}{Yujie Wei}, \bibinfo{person}{Tao Feng}, \bibinfo{person}{Yining Pan}, \bibinfo{person}{Yingya Zhang}, \bibinfo{person}{Ziwei Liu}, \bibinfo{person}{Samuel Albanie}, {and} \bibinfo{person}{Dong Ni}.} \bibinfo{year}{2024}\natexlab{}.
\newblock \showarticletitle{InstructVideo: instructing video diffusion models with human feedback}. In \bibinfo{booktitle}{\emph{Proc. CVPR}}. \bibinfo{pages}{6463--6474}.
\newblock


\bibitem[Zhang et~al\mbox{.}(2024b)]%
        {zhang2024show}
\bibfield{author}{\bibinfo{person}{David~Junhao Zhang}, \bibinfo{person}{Jay~Zhangjie Wu}, \bibinfo{person}{Jia-Wei Liu}, \bibinfo{person}{Rui Zhao}, \bibinfo{person}{Lingmin Ran}, \bibinfo{person}{Yuchao Gu}, \bibinfo{person}{Difei Gao}, {and} \bibinfo{person}{Mike~Zheng Shou}.} \bibinfo{year}{2024}\natexlab{b}.
\newblock \showarticletitle{Show-1: Marrying pixel and latent diffusion models for text-to-video generation}.
\newblock \bibinfo{journal}{\emph{Int. J. Comput. Vis.}} (\bibinfo{year}{2024}), \bibinfo{pages}{1--15}.
\newblock


\bibitem[Zhang et~al\mbox{.}(2023)]%
        {zhang2023adding}
\bibfield{author}{\bibinfo{person}{Lvmin Zhang}, \bibinfo{person}{Anyi Rao}, {and} \bibinfo{person}{Maneesh Agrawala}.} \bibinfo{year}{2023}\natexlab{}.
\newblock \showarticletitle{Adding conditional control to text-to-image diffusion models}. In \bibinfo{booktitle}{\emph{Proc. ICCV}}. \bibinfo{pages}{3836--3847}.
\newblock


\bibitem[Zhang et~al\mbox{.}(2024a)]%
        {zhang2024mimicmotion}
\bibfield{author}{\bibinfo{person}{Yuang Zhang}, \bibinfo{person}{Jiaxi Gu}, \bibinfo{person}{Li-Wen Wang}, \bibinfo{person}{Han Wang}, \bibinfo{person}{Junqi Cheng}, \bibinfo{person}{Yuefeng Zhu}, {and} \bibinfo{person}{Fangyuan Zou}.} \bibinfo{year}{2024}\natexlab{a}.
\newblock \showarticletitle{Mimicmotion: High-quality human motion video generation with confidence-aware pose guidance}.
\newblock \bibinfo{journal}{\emph{arXiv preprint arXiv:2406.19680}} (\bibinfo{year}{2024}).
\newblock


\bibitem[Zhao et~al\mbox{.}(2023)]%
        {zhao2023motiondirector}
\bibfield{author}{\bibinfo{person}{Rui Zhao}, \bibinfo{person}{Yuchao Gu}, \bibinfo{person}{Jay~Zhangjie Wu}, \bibinfo{person}{David~Junhao Zhang}, \bibinfo{person}{Jiawei Liu}, \bibinfo{person}{Weijia Wu}, \bibinfo{person}{Jussi Keppo}, {and} \bibinfo{person}{Mike~Zheng Shou}.} \bibinfo{year}{2023}\natexlab{}.
\newblock \bibinfo{title}{MotionDirector: Motion Customization of Text-to-Video Diffusion Models}.
\newblock
\newblock
\showeprint[arxiv]{2310.08465}~[cs.CV]


\bibitem[Zhou et~al\mbox{.}(2024)]%
        {zhou2024allegro}
\bibfield{author}{\bibinfo{person}{Yuan Zhou}, \bibinfo{person}{Qiuyue Wang}, \bibinfo{person}{Yuxuan Cai}, {and} \bibinfo{person}{Huan Yang}.} \bibinfo{year}{2024}\natexlab{}.
\newblock \showarticletitle{Allegro: Open the Black Box of Commercial-Level Video Generation Model}.
\newblock \bibinfo{journal}{\emph{arXiv preprint arXiv:2410.15458}} (\bibinfo{year}{2024}).
\newblock


\bibitem[Zhu et~al\mbox{.}(2024)]%
        {zhu2024champ}
\bibfield{author}{\bibinfo{person}{Shenhao Zhu}, \bibinfo{person}{Junming~Leo Chen}, \bibinfo{person}{Zuozhuo Dai}, \bibinfo{person}{Yinghui Xu}, \bibinfo{person}{Xun Cao}, \bibinfo{person}{Yao Yao}, \bibinfo{person}{Hao Zhu}, {and} \bibinfo{person}{Siyu Zhu}.} \bibinfo{year}{2024}\natexlab{}.
\newblock \showarticletitle{Champ: Controllable and Consistent Human Image Animation with 3D Parametric Guidance}. In \bibinfo{booktitle}{\emph{EECV}}.
\newblock


\end{thebibliography}

\end{document}